\newif\iflongversion
\longversionfalse
\longversiontrue

\documentclass[runningheads]{llncs}
\usepackage[T1]{fontenc}
\usepackage[breaklinks, colorlinks, linkcolor=blue, citecolor=blue, urlcolor=black]{hyperref}
\usepackage{graphicx}
\usepackage{amsmath}
\usepackage{amssymb}
\usepackage{color}
\usepackage{paralist}
\usepackage{bbding}

\usepackage{color}

\usepackage{tikz}
\usetikzlibrary{positioning, intersections, calc, patterns.meta, shapes.geometric, backgrounds, fit}

\usepackage{ifthen}

\usepackage{ifthen}

\provideboolean{detectedSTOC}
\provideboolean{detectedFOCS}
\provideboolean{detectedElsevier}
\provideboolean{detectedNOW}
\provideboolean{detectedLMCS}
\provideboolean{detectedIEEE}
\provideboolean{detectedPoster}
\provideboolean{detectedSIAM}
\provideboolean{detectedLNCS}
\provideboolean{detectedACM}
\provideboolean{detectedACMconf}
\provideboolean{detectedSigplanconf}
\provideboolean{detectedToC}
\provideboolean{detectedLIPIcs}
\provideboolean{detectedAAAI}
\provideboolean{detectedIJCAI}
\provideboolean{detectedCompCplx}
\provideboolean{detectedEasyChair}
\provideboolean{detectedArticle}
\provideboolean{detectedReport}
\provideboolean{detectedThesis}

\makeatletter

\@ifclassloaded{sig-alternate}
{\setboolean{detectedSTOC}{true}}
{\setboolean{detectedSTOC}{false}}

\@ifclassloaded{elsarticle}
{\setboolean{detectedElsevier}{true}}
{\setboolean{detectedElsevier}{false}}

\@ifclassloaded{now}
{\setboolean{detectedNOW}{true}}
{\setboolean{detectedNOW}{false}}

\@ifclassloaded{lmcs}
{\setboolean{detectedLMCS}{true}}
{\setboolean{detectedLMCS}{false}}

\@ifclassloaded{IEEEtran} {
  \setboolean{detectedIEEE}{true}
  \ifCLASSOPTIONconference {
    \setboolean{detectedFOCS}{true}
  }
  \else {
    \setboolean{detectedFOCS}{false}
  }
  \fi
}
{
  \setboolean{detectedFOCS}{false}
  \setboolean{detectedIEEE}{false}
}

\@ifclassloaded{siamart171218}
{\setboolean{detectedSIAM}{true}}
{\setboolean{detectedSIAM}{false}}

\@ifclassloaded{llncs}
{\setboolean{detectedLNCS}{true}}
{\setboolean{detectedLNCS}{false}}

\@ifclassloaded{acmsmall}
{\setboolean{detectedACM}{true}}
{\setboolean{detectedACM}{false}}

\@ifclassloaded{acmart}
{\setboolean{detectedACMconf}{true}}
{\setboolean{detectedACMconf}{false}}

\@ifclassloaded{sigplanconf}
{\setboolean{detectedSigplanconf}{true}}
{\setboolean{detectedSigplanconf}{false}}

\@ifclassloaded{toc}
{\setboolean{detectedToC}{true}}
{\setboolean{detectedToC}{false}}

\@ifclassloaded{lipics}
{\setboolean{detectedLIPIcs}{true}}
{\@ifclassloaded{lipics-v2019}
  {\setboolean{detectedLIPIcs}{true}}
  {\@ifclassloaded{oasics-v2019}
    {\setboolean{detectedLIPIcs}{true}}
    {\setboolean{detectedLIPIcs}{false}}
  }
}

\@ifclassloaded{cc}
{\setboolean{detectedCompCplx}{true}}
{\setboolean{detectedCompCplx}{false}}

\@ifclassloaded{easychair}
{\setboolean{detectedEasyChair}{true}}
{\setboolean{detectedEasyChair}{false}}

\@ifpackageloaded{aaai}
{\setboolean{detectedAAAI}{true}}        
{\@ifpackageloaded{aaai18}
  {\setboolean{detectedAAAI}{true}}
  {\@ifpackageloaded{aaai20}       
    {\setboolean{detectedAAAI}{true}}
    {\setboolean{detectedAAAI}{false}}}}

\@ifpackageloaded{ijcai18}
{\setboolean{detectedIJCAI}{true}}        
{\@ifpackageloaded{ijcai19}
  {\setboolean{detectedIJCAI}{true}}        
  {\setboolean{detectedIJCAI}{false}}}

\@ifclassloaded{sciposter}
{\setboolean{detectedPoster}{true}}
{\setboolean{detectedPoster}{false}}

\@ifclassloaded{article}
{\setboolean{detectedArticle}{true}}
{\setboolean{detectedArticle}{false}}

\makeatother

\ifthenelse{\not \isundefined{\examen} 
  \and \not \isundefined{\disputationsdatum} 
  \and \not \isundefined{\disputationslokal}}   
  {\setboolean{detectedThesis}{true}}
  {\setboolean{detectedThesis}{false}}

\ifthenelse{\boolean{detectedArticle} \or \boolean{detectedThesis}
  \or \boolean{detectedSTOC} \or \boolean{detectedFOCS}
  \or \boolean{detectedSIAM} \or \boolean{detectedIEEE}
  \or \boolean{detectedPoster}}
{\setboolean{detectedReport}{false}}
{\setboolean{detectedReport}{true}}

\usepackage{xspace}
\ifthenelse
{\boolean{detectedElsevier} \or \boolean{detectedSIAM} 
  \or \boolean{detectedLIPIcs}}
{}
{\usepackage{varioref}}

\makeatletter
\AtBeginDocument{%
  \def\doi#1{\url{https://doi.org/#1}}}
\makeatother

\DeclareMathAlphabet{\mathsfsl}{OT1}{cmss}{m}{sl}

\DeclareRobustCommand{\BibTeX}{%
  {\normalfont B\kern-.05em{\scshape i\kern-.025em b}\kern-.08em \TeX}%
}

\ifthenelse{\boolean{detectedToC}}{}
{

}

\newcommand{\ceiling}[1]{\lceil #1 \rceil}

\newcommand{\MAXOFEXPR}[2][]{\max_{#1} \left\{ #2 \right\}}
\newcommand{\MINOFEXPR}[2][]{\min_{#1} \left\{ #2 \right\}}
\newcommand{\Maxofexpr}[2][]{\max_{#1} \bigl\{ #2 \bigr\}}
\newcommand{\Minofexpr}[2][]{\min_{#1} \bigl\{ #2 \bigr\}}

\newcommand{\MAXOFSET}[3][:]%
     {\ifthenelse{\equal{#1}{;}}%
     {\MAXOFEXPR{ #2 \,;\, #3 }}
     {\ifthenelse{\equal{#1}{:}}%
     {\MAXOFEXPR{ #2 \,:\, #3 }}
     {\max \twincommandJN{\left\{}{#2}{\left#1}{\right}{\,#3}{\right\}}}}}
\newcommand{\MINOFSET}[3][:]%
     {\ifthenelse{\equal{#1}{;}}%
     {\MINOFEXPR{ #2 \,;\, #3 }}
     {\ifthenelse{\equal{#1}{:}}%
     {\MINOFEXPR{ #2 \,:\, #3 }}
     {\min \twincommandJN{\left\{}{#2}{\left#1}{\right}{\,#3}{\right\}}}}}

\newcommand{\Maxofset}[3][:]%
     {\ifthenelse{\equal{#1}{;}}%
     {\Maxofexpr{ #2 \,;\, #3 }}
     {\ifthenelse{\equal{#1}{:}}%
     {\Maxofexpr{ #2 \,:\, #3 }}
     {\max \twincommandJN{\bigl\{}{#2}{\bigl#1}{\bigr}{\,#3}{\bigr\}}}}}
\newcommand{\Minofset}[3][:]%
     {\ifthenelse{\equal{#1}{;}}%
     {\Minofexpr{ #2 \,;\, #3 }}
     {\ifthenelse{\equal{#1}{:}}%
     {\Minofexpr{ #2 \,:\, #3 }}
     {\min \twincommandJN{\bigl\{}{#2}{\bigl#1}{\bigr}{\,#3}{\bigr\}}}}}

\DeclareMathOperator{\Expop}{E}

\ifthenelse{\boolean{detectedLMCS}}
{}
{}

\newcommand{\twincommandJN}[6]%
    {#1#2#3\vphantom{#2#5}\mspace{-2.05mu}#4.#5#6}

\newcommand{\CondExp}[2]%
    {\Expop\twincommandJN{\bigl[}{#1}{\bigl|}{\bigr}{\,#2}{\bigr]}}
\newcommand{\CONDEXP}[2]%
     {\Expop\twincommandJN{\left[}{#1}{\left|}{\right}{\,#2}{\right]}}

\newcommand{\Condprob}[3][]%
    {\Pr_{#1}\twincommandJN{\bigl[}{#2}{\bigl|}{\bigr}{\,#3}{\bigr]}}
\newcommand{\CONDPROB}[3][]%
    {\Pr_{#1}\twincommandJN{\left[}{#2}{\left|}{\right}{\,#3}{\right]}}

\newcommand{\set}[1]{\{ #1 \}}
\newcommand{\Set}[1]{\bigl\{ #1 \bigr\}}

\newcommand{\setdescr}[3][\mid]{\set{ #2 #1 #3 }}
\newcommand{\Setdescr}[3][|]%
     {\ifthenelse{\equal{#1}{;}}%
     {\Set{ #2 \,;\, #3 }}
     {\ifthenelse{\equal{#1}{:}}%
     {\Set{ #2 \,:\, #3 }}
     {\twincommandJN{\bigl\{}{#2\,}{\bigl#1}{\bigr}{\,#3}{\bigr\}}}}}
\newcommand{\SETDESCR}[3][|]%
     {\twincommandJN{\left\{}{#2\,}{\left#1}{\right}{\,#3}{\right\}}}

\newcommand{\Setdescrbrackets}[3][|]%
     {\twincommandJN{\bigl[}{#2}{\bigl#1}{\bigr}{\,#3}{\bigr]}}
\newcommand{\SETDESCRBRACKETS}[3][|]%
     {\twincommandJN{\left[}{#2}{\left#1}{\right}{\,#3}{\right]}}

\newcommand{\setsize}[1]{\lvert#1\rvert}

\newcommand{\olnot}[1]{\overline{#1}}

\newcommand{\nvar}{n}

\newcommand{\nclause}{m}

\newcommand{\clwidth}{k}

\newcommand{\randkcnfnclwrepl}[3][\clwidth]%
        {\ensuremath{\mathcal{F}^{#2, #3}_{#1}}}
\newcommand{\randkcnfnclwreplstd}%
        {\randkcnfnclwrepl{\clwidth}{\nvar}{\nclause}}

\newcommand{\complclassformat}[1]%
        {\textrm{\upshape{\textsf{#1}}}\xspace}
\newcommand{\cocomplclass}[1]%
        {\textrm{\upshape{\textsf{co#1}}}\xspace}

\newcommand{\DTIMEadviceclass}[2]%
    {\ensuremath{\complclassformat{DTIME}\bigl(#1\bigr)/{#2}}}

\newcommand{\NP}{\complclassformat{NP}}

\newcommand{\PCPalph}[5]%
    {\ensuremath{\complclassformat{PCP}_{{#1},{#2}}[{#3}, {#4}, {#5}]}}
\newcommand{\PCP}[4]%
    {\ensuremath{\complclassformat{PCP}_{{#1},{#2}}[{#3}, {#4}]}}

\newcommand{\eqcomma}{\enspace ,}

\renewcommand{\eqcomma}{\, ,}

\ifthenelse{\boolean{detectedToC}}{}
  { 
    }
\ifthenelse{\boolean{detectedToC}}{}{
\newcommand{\ie}{i.e.,\ }

}

\ifthenelse{\boolean{detectedLIPIcs} \or \boolean{detectedIJCAI}}
{\renewcommand{\st}{\errmessage{Please do not use st}}}
{\ifthenelse{\isundefined{\st}}
  {\newcommand{\st}{such that\xspace}}
  {}}      

\ifthenelse{\boolean{detectedIEEE}}{}{}

\ifthenelse
{\isundefined{\refeq}}
{\newcommand{\refeq}[1]{\eqref{#1}}}
{\renewcommand{\refeq}[1]{\eqref{#1}}}

\newcommand{\derives}{\vdash}

\newcommand{\varx}{\ensuremath{x}}
\ifthenelse{\boolean{detectedACMconf}}
{}
{}

\ifthenelse{\boolean{detectedACMconf}}
{
 }
{
 }

\newcommand{\litl}{\ensuremath{\ell}}

\newcommand{\SETSOFVARSORLIT}[2]%
        {\mathit{#1}\left({#2}\right)}
\newcommand{\setsofvarsorlit}[2]%
        {\mathit{#1}({#2})}
\newcommand{\Setsofvarsorlit}[2]%
        {\mathit{#1}\bigl({#2}\bigr)}

\newcommand{\restrict}[2]{{{#1}\!\!\upharpoonright_{#2}}}

\newcommand{\derivabbrev}[2]{\bigl( #1 \vdash #2 \bigr)}
\newcommand{\derivabbrevsmall}[2]{( #1 \vdash #2 )}
\newcommand{\derivabbrevcompact}[2]{\bigl( #1 \vdash #2 \bigr)}

\newcommand{\refutabbrevsmall}[1]{\derivabbrevsmall{#1}{\!\bot}}
\newcommand{\refutabbrevcompact}[1]{\derivabbrevcompact{#1}{\!\bot}}

\newcommand{\genericrefsmall}[3]%
    {{\mathit{#1}}_{#2}\refutabbrevsmall{#3}}
\newcommand{\genericrefcompact}[3]%
    {{\mathit{#1}}_{#2}\refutabbrevcompact{#3}}
\newcommand{\genericderiv}[4]%
    {{\mathit{#1}}_{#2}\derivabbrev{#3}{#4}}
\newcommand{\genericderivsmall}[4]%
    {{\mathit{#1}}_{#2}\derivabbrevsmall{#3}{#4}}
\newcommand{\genericderivcompact}[4]%
    {{\mathit{#1}}_{#2}\derivabbrevcompact{#3}{#4}}
\newcommand{\generictaut}[3]%
    {{\mathit{#1}}_{#2}\derivabbrev{}{#3}}
\newcommand{\generictautcompact}[3]%
    {{\mathit{#1}}_{#2}\derivabbrevcompact{}{#3}}
\newcommand{\generictautsmall}[3]%
    {{\mathit{#1}}_{#2}\derivabbrevsmall{}{#3}}

\newcommand{\formulaformat}[1]{\mathit{#1}}

\newcommand{\extendedversion}[1]{\widetilde{#1}}

\newcommand{\epopnot}[1]%
    {\extendedversion{\formulaformat{POP}}_{#1}}

\newcommand{\elopnot}[1]%
    {\extendedversion{\formulaformat{LOP}}_{#1}}

\newcommand{\ephpnot}[2]%
    {\vphantom{\extendedversion{\formulaformat{PHP}}}
      {\smash{\extendedversion{\formulaformat{PHP}}}
        \vphantom{\formulaformat{PHP}}}^{#1}_{#2}}

\newcommand{\efphpnot}[2]%
    {\vphantom{\extendedversion{\formulaformat{FPHP}}}
      {\smash{\extendedversion{\formulaformat{FPHP}}}
        \vphantom{\formulaformat{FPHP}}}^{#1}_{#2}}

\newcommand{\ontophpnot}[2]%
    {\formulaformat{Onto}\text{-}\formulaformat{PHP}^{#1}_{#2}}
\newcommand{\ontofphpnot}[2]%
    {\formulaformat{Onto}\text{-}\formulaformat{FPHP}^{#1}_{#2}}

\newcommand{\graphontophpnot}[1][G]%
    {\text{$\formulaformat{Onto}$-$\formulaformat{PHP}$}({#1})}

\newcommand{\perfectmatchingnot}[1][G]%
    {\formulaformat{PM}({#1})}
\newcommand{\wcnfinstance}{\mathcal{F}^W} 
\newcommand{\instance}{\mathcal{F}} 
\newcommand{\opt}{\textsc{opt}}

\newcommand{\up}{\vdash_{_{_\textsc{up}}}}
\newcommand{\confl}{\bot}
\newtheorem{observation}{Observation}

\newcommand{\assmtrho}{\rho}
\newcommand{\witness}{\omega}

\newcommand{\core}{\mathcal{C}}
\newcommand{\derived}{\mathcal{D}}
\newcommand{\formula}{F} 
\newcommand{\hards}{\formula_H} 
 \newcommand{\softs}{\formula_S} 

\newcommand{\weight}{w} 
\newcommand{\cost}{\textsc{cost}} 

\newcommand{\objectivetransform}{\textsc{ObjMaxSAT}} 
\newcommand{\asPB}{\textsc{PB}} 

\newcommand{\obj}{O}
\newcommand{\obc}{W} 

\newcommand{\objorig}{\obj}
\newcommand{\objmodified}{\obj^\prime}

\newcommand{\zeroone}{\mbox{$0$--$1$}\xspace}

\newcommand{\toolname}[1]{\textsc{#1}\xspace}
\newcommand{\veripb}{\toolname{VeriPB}}
\newcommand{\CakePB}{\toolname{CakePB}}
\newcommand{\cakepb}{\CakePB}
\newcommand{\cakepbwcnf}{\toolname{CakePB\-wcnf}}
\newcommand{\maxpre}{\toolname{MaxPre}}

\newcommand{\proofsystem}[1]{\toolname{#1}}
\newcommand{\tracecheck}{\proofsystem{TraceCheck}}
\newcommand{\rupname}{\proofsystem{RUP}}

\newcommand{\drat}{\proofsystem{DRAT}}
\newcommand{\grit}{\proofsystem{GRIT}}
\newcommand{\lrat}{\proofsystem{LRAT}}

\newcommand{\HOL}{HOL4\xspace}
\renewcommand{\HOL}{\toolname{HOL4}}

\newcommand{\CakeML}{\toolname{CakeML}}

\usepackage{holtexbasic}
\usepackage{environ}
\NewEnviron{holthmenv}{%
  \scalebox{1.0}{\begin{array}[t]{l}
  \BODY
  \end{array}}}

\renewcommand{\HOLTokenTurnstile}{\ensuremath{\vdash\!\!}}
\renewcommand{\HOLConst}[1]{\textsf{\small #1}}

\renewcommand{\HOLSymConst}[1]{\HOLConst{#1}}

\renewcommand{\HOLKeyword}[1]{\ensuremath{\mathsf{#1}}}
\renewcommand{\HOLTokenBar}{\ensuremath{\mathtt{|}}}

\renewcommand{\HOLStringLitDG}[1]{\scalebox{0.9}{\texttt{"#1"}}}

\newcommand{\TODO}[1]{\textcolor{red}{#1}}

\makeatletter
\def\orcidID#1{\href{http://orcid.org/#1}{\protect\raisebox{-1.25pt}{\protect\includegraphics{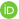}}}}
\makeatother

\usepackage{cleveref}
\crefname{table}{Table}{Tables}
\Crefname{table}{Table}{Tables}
\crefname{figure}{Figure}{Figures}
\Crefname{figure}{Figure}{Figures}
\crefname{section}{Section}{Sections}
\Crefname{section}{Section}{Sections}

\usepackage{makecell}

\begin{document}

\title{Certified MaxSAT Preprocessing \iflongversion \\ \emph{Extended version including appendix} \fi}
%
%
\author{Hannes Ihalainen\inst{1}\Envelope\orcidID{0000-0002-4608-7549} \and
  Andy Oertel\inst{2,3}\orcidID{0000-0001-9783-6768} \and
  Yong Kiam Tan\inst{4}\orcidID{0000-0001-7033-2463} \and
  Jeremias Berg\inst{1}\orcidID{0000-0001-7660-8061} \and
  Matti J\"arvisalo\inst{1}\orcidID{0000-0003-2572-063X} \and
  Magnus O. Myreen\inst{5}\orcidID{0000-0002-9504-4107} \and
  Jakob Nordstr\"om\inst{3,2}\orcidID{0000-0002-2700-4285} 
}
\authorrunning{H. Ihalainen et al.}
%
\institute{Department of Computer Science, University of Helsinki, Helsinki, Finland\\
\email{\{hannes.ihalainen,jeremias.berg,matti.jarvisalo\}@helsinki.fi}
\and
Lund University, Lund, Sweden\\
  \email{andy.oertel@cs.lth.se
  }
  \and
  University of Copenhagen, Copenhagen, Denmark\\
  \email{jn@di.ku.dk}
  \and
  Institute for Infocomm Research (I$^2$R), A*STAR, Singapore\\
  \email{tanyk1@i2r.a-star.edu.sg}
  \and
  Chalmers University of Technology, Gothenburg, Sweden\\
  \email{myreen@chalmers.se}
}
\maketitle              

\begin{abstract}

  Building on the progress in Boolean satisfiability (SAT) solving over the
  last decades, maximum satisfiability (MaxSAT) has become a viable approach
  for solving \NP-hard optimization problems,
  but ensuring correctness of MaxSAT solvers has remained an important concern.
  For SAT, this is largely a solved problem thanks to the use of proof logging,
  meaning that solvers emit machine-verifiable proofs of (un)satisfiability to
  certify correctness. However, for MaxSAT, proof logging solvers have started
  being developed only very recently.
  Moreover, these nascent efforts have only targeted the core solving process,
  ignoring the preprocessing phase where input problem instances can be
  substantially reformulated before being passed on to the solver proper.

  In this work, we demonstrate how pseudo-Boolean proof logging can be
  used to certify the correctness of a wide range of modern MaxSAT
  preprocessing techniques.
  By combining and extending the \veripb and \cakepb tools, we provide
  formally verified, end-to-end proof checking that the input and
  preprocessed output MaxSAT problem instances have the same optimal value.
  An extensive evaluation on applied MaxSAT benchmarks shows that our
  approach is feasible in practice.

%

\end{abstract}


\newcommand{\subsectionintro}[1]{\subsection{#1}}
\renewcommand{\subsectionintro}[1]{\noindent\textbf{#1.}}
\renewcommand{\subsectionintro}[1]{\paragraph{#1}}
\renewcommand{\subsectionintro}[1]{\subsection{#1}}

\section{Introduction}
\label{sec:intro}

The development of Boolean satisfiability (SAT) solvers is arguably
one of the true success stories of modern computer science---today,
SAT solvers are routinely used as core
engines in many types
of complex automated reasoning systems.
One example
of this is SAT-based optimization, usually referred to as
\emph{maximum satisfiability (MaxSAT) solving.}
The improved performance of SAT
solvers, coupled with increasingly sophisticated techniques for using
SAT solver calls to reason about optimization problems, have made
MaxSAT solvers a powerful tool
for tackling real-world \NP-hard
optimization problems~\cite{HandbookSAT21}.

Modern MaxSAT solvers are quite intricate pieces of software, however,
and it
has been shown repeatedly in the MaxSAT
evaluations~\cite{MaxSATevaluations} that even
the best  
solvers sometimes report incorrect results.  This was previously a
serious issue also for SAT solvers (see,
e.g.,~\cite{BLB10AutomatedTesting}), but the SAT community has
essentially eliminated this problem by
requiring  
that solvers should be
\emph{certifying}~\cite{ABMRS11IntroCertifyingAlgo,MMNS11CertifyingAlgorithms},
\ie
not only report whether a given formula is
satisfiable or unsatisfiable but also produce a machine-verifiable
proof
that this conclusion is correct.
Many different SAT proof formats such as
\rupname~\cite{GN03Verification},
\tracecheck~\cite{Biere06TraceCheck},
\grit~\cite{CMS17EfficientCertified}, and
\lrat~\cite{CHHKS17EfficientCertified}
have been proposed,
with
\drat~\cite{HHW13Trimming,HHW13Verifying,WHH14DRAT}
established as the de facto standard;
for the last ten years, proof logging has
been compulsory in the (main track of the) SAT
competitions~\cite{SATcompetition}.
It is all the more striking, then, that until
recently no similar
developments have been observed in MaxSAT solving.

\subsectionintro{Previous Work}
A first natural question to ask---since MaxSAT solvers are based on
repeated calls to
SAT solvers---is whether SAT proof logging techniques could be used for
MaxSAT.
Although clausal proof systems such as \drat
could in
principle
certify
MaxSAT reasoning, the
overhead of translating 
claims about the cost of solutions 
into disjunctive clauses 
that \drat reasons with
seems
too large to be feasible in practice.

While there have been
several
attempts to design proof logging
methods specifically for MaxSAT solving~%
\cite{DBLP:journals/ai/BonetLM07,FMSV20MaxSAT,IBJ22ClauseRedundancy,DBLP:journals/jar/LarrosaNOR11,MIBMB19DRMaxSAT,DBLP:conf/ictai/MorgadoM11,PCH20TowardsBridging,PCH21ProofBuilder,PCH22ProofsCertificatesMaxSAT},
none of these has come close to providing a general proof logging
solution, because they apply only for very specific algorithm
implementations and/or fail to capture the full range of techniques
used.  Recent papers have instead proposed using pseudo-Boolean
proof logging with
\veripb~\cite{BGMN23Dominance,GN21CertifyingParity}
to certify correctness of so-called
solution-improving solvers~\cite{VWB22QMaxSATpb}
and core-guided solvers~\cite{BBNOV23CertifiedCoreGuided}.
Although these works demonstrate, for the first time, practical proof logging
for modern MaxSAT solving, the methods developed thus far only apply to the
core solving process. This ignores the MaxSAT preprocessing phase, where the
input formula can undergo major
reformulation. State-of-the-art
solvers sometimes use
stand-alone preprocessor tools, or sometimes integrate
preprocessing-style reasoning more tightly within the MaxSAT solver
engine, 
to speed up the search for optimal solutions.
Some of these preprocessing techniques
are lifted from SAT to
MaxSAT, but there are also native MaxSAT preprocessing methods that
reason about the objective function being optimized and therefore lack
analogies in SAT solving.

\subsectionintro{Our Contribution}
In this paper, we show, for the first time, how to use pseudo-Boolean proof
logging with \veripb to produce proofs of correctness for a wide range
of preprocessing techniques used in modern MaxSAT
solvers.
\veripb proof logging has previously been successfully used not only
for core MaxSAT search as discussed above, but also for
advanced SAT solving techniques (including symmetry breaking)~%
\cite{BGMN23Dominance,GMNO22CertifiedCNFencodingPB,GN21CertifyingParity},
subgraph solving~%
\cite{GMMNPT20CertifyingSolvers,GMMNOT24Subgraph,GMN20SubgraphIso},
constraint programming~%
\cite{EGMN20Justifying,GMN22AuditableCP,MM23ProofLogging,MMN24ProofLoggingCircuitConstraint},
and \zeroone ILP  presolving~\cite{HOGN24CertifyingMIPpresolve},
and we add MaxSAT preprocessing to this list.

In order to do so, we extend the \veripb proof format to include an output
section where a reformulated output can be presented, and where the
pseudo-Boolean proof establishes that this output formula and the input
formula are \emph{equioptimal}, \ie have optimal solutions of the same value.
We also enhance \cakepb~\cite{BMMNOT23DocumentationVeriPB,GMMNOT24Subgraph}---a
verified proof checker for pseudo-Boolean
proofs---to handle proofs of
reformulation. In this way, we obtain a formally verified toolchain for
certified preprocessing of MaxSAT instances.

It is worth noting that, although preprocessing is also a critical component in
SAT solving, we are not aware of any proof logging tool that supports proofs of
correctness even for reformulated decision problems, meaning that formulas are
\emph{equisatisfiable}---the \drat format and tools support proofs that
satisfiability of an input CNF formula $F$ implies satisfiability of an output
CNF formula $G$ but not the converse direction (except in the special case
where $F$ is a subset of $G$).
To the best of our knowledge, our work presents the first practical tool for
proving (two-way) equisatisfiability or equioptimality of reformulated
problems.

We have performed computational experiments running a MaxSAT
preprocessor with proof logging and proof checking on benchmarks from
the MaxSAT evaluations.  Although there is certainly room for
improvements in performance, these experiments provide empirical
evidence for the feasibility of certified preprocessing for real-world MaxSAT benchmarks.


\subsectionintro{Organization of This Paper}
After reviewing
preliminaries in 
\cref{sec:preliminaries},
we explain our
pseudo-Boolean proof logging for MaxSAT
preprocessing in
\cref{sec:logging-techniques},
and
\cref{sec:verified}
discusses verified proof checking.
We present results from a computational evaluation in 
\cref{sec:experiments},
after which we conclude with a summary and outlook for future work in
\cref{sec:conclusion}.


\section{Preliminaries}
\label{sec:preliminaries}

We write~$\varx$ to denote a Boolean variable, which we think of as
$\set{0,1}$-valued, and write
$\olnot{\varx}$ as a shorthand \mbox{for $1-x$}.
We use~$\litl$ to denote such \emph{positive} and
\emph{negative literals}, respectively.
A \emph{clause}
$C=\litl_1\lor \litl_2\lor\dots \lor \litl_k$
is a disjunction of literals and a
formula in
\emph{conjunctive normal form (CNF)}
$\formula = C_1 \land \ldots \land C_m$
is a conjunction of clauses,
where we think of clauses and formulas as sets when convenient,
so that there are no repetitions and order is irrelevant.

A
\emph{pseudo-Boolean (PB) constraint}
is a \zeroone
linear inequality
\mbox{$\sum_j a_j \litl_j \geq b$},
where, when convenient, we can assume all literals~$\litl_j$ to refer to
distinct variables and all integers~$a_j$ and~$b$ to be
non-negative
(so-called \emph{normalized form}),
and a
\emph{pseudo-Boolean formula} is
a
conjunction of such constraints.
We can identify the clause
$C \equiv \litl_1\lor \litl_2\lor\dots \lor \litl_k$
with the pseudo-Boolean constraint
$\asPB(C) = \litl_1 +  \litl_2 + \dots +  \litl_k \geq 1$,
so a CNF formula $\formula$ is just
a special type of PB formula
$\asPB(\formula) = \{ \asPB(C) \mid C \in \formula\}$.

A \emph{(partial) assignment}~$\assmtrho$
mapping variables to~$\set{0,1}$,
and extending to literals by
$\assmtrho(\olnot{\varx}) = 1 - \assmtrho(\varx)$
for $\varx$ in the domain of~$\assmtrho$,
satisfies
a  
PB constraint
\mbox{$\sum_j a_j \litl_j \geq b$}
if
$\sum_{\litl_j : \assmtrho(\litl_j) = 1} a_j  \geq b$
(assuming normalized form),
and a PB formula is satisfied by~$\assmtrho$ if all constraints in it are.
We also refer to total satisfying assignments~$\assmtrho$ as \emph{solutions}.
In a
\emph{pseudo-Boolean optimization (PBO)} problem
we ask for the solution minimizing an \emph{objective} function
\mbox{$\obj = \sum_j c_j \litl_j  + \obc$},
where $c_j$ and $\obc$ are integers and $\obc$ represents a trivial lower bound on the minimum cost.


\subsection{Pseudo-Boolean Proof Logging
  Using Cutting Planes
}
\label{subsec:prelim-veripb}

Pseudo-Boolean proof logging---as supported in \veripb---is based on the
\emph{cutting planes} proof system~\cite{CCT87ComplexityCP}
with extensions as discussed briefly below. We refer
the reader
to~\cite{BN21ProofCplxSATplusCrossref} for more information about
cutting planes and
to~\cite{BGMN23Dominance,GN21CertifyingParity,HOGN24CertifyingMIPpresolve}
for detailed information about the  \veripb proof system and format.

A pseudo-Boolean proof maintains
two sets of
\emph{core constraints}~$\core$
and
\emph{derived constraints}~$\derived$
under which the objective~$\obj$ should be minimized.
At the start of the proof,
$\core$~is  initialized to the constraints in the input formula.
Any constraints derived by the rules described below are  
placed in~$\derived$, 
from where they can later be
moved to~$\core$ (but not vice versa).
Proofs maintain the invariant that the optimal value of any solution
to~$\core$
and to the original input problem is the same.
Hence, deletions from $\core$ cannot be done arbitrarily but require the so called checked deletion rule discussed later.
New constraints can be derived from
$\core \cup  \derived$
by performing \emph{addition} of two constraints
or \emph{multiplication}
of a constraint by a positive integer,
and
\emph{literal axioms} $\ell \geq 0$
can be used at any time.
Additionally, we can apply \emph{division} to
$\sum_{j} a_j \ell_j \geq b$
by a positive integer~$d$ followed by rounding up to obtain
$\sum_{j} \ceiling{a_j / d} \ell_j \geq \ceiling{b / d}$,
and \emph{saturation} to yield
$\sum_j \min \set{a_j, b} \cdot \ell_j \geq b$
(where we again assume normalized form).

The negation of a constraint
$C
=  
\sum_{j} a_j \ell_j \geq b$
is
$
\neg C
= 
\sum_{j} a_j \ell_j \leq b - 1$.
For a (partial) assignment~$\rho$
we write
$\restrict{C}{\rho}$
for the \emph{restricted constraint} obtained by replacing variables
assigned by~$\rho$ with their values and simplifying.
We say that $C$
\emph{unit propagates $\ell$ under $\rho$} if $\restrict{C}{\rho}$
cannot be satisfied unless $\ell$ is assigned to~$1$.
If repeated unit propagation on all constraints in
$\core \cup \derived \cup \set{\neg C}$,
starting with the empty assignment
$\rho= \emptyset$,
leads to contradiction in the form of an unsatisfiable constraint,
we say that
$C$ follows by \emph{reverse unit propagation (RUP)} from
$\core \cup \derived$.
Such (efficiently verifiable) RUP steps are allowed in
proofs as a convenient way to
avoid writing out an explicit cutting planes derivation.
We use the same notation
$\restrict{C}{\witness}$
to denote the result of applying to~$C$ a
\emph{(partial) substitution}~$\witness$, which can also map variables to
other literals in addition to~$0$ and~$1$,
and extend this notation to sets in the obvious way by the set image of
$\restrict{(\cdot)}{\witness}$.

In addition to the above rules, which derive
semantically implied constraints, there is a
\emph{redundance-based strengthening rule},
or just \emph{redundance rule} for short,
that can derive a
non-implied constraint~$C$ as long as this does not change the feasibility
or optimal value.
This can be guaranteed by considering a \emph{witness substitution}~$\witness$ such
that for any total assignment $\alpha$ satisfying
$\core \cup \derived$ but violating~$C$, the composition
$\alpha \circ \witness$
is another total assignment that satisfies
$\core \cup \derived \cup \set{C}$
and yields
an objective value that is at least as good for the objective~$\obj$.
Formally, $C$ can be derived from $\core \cup \derived$
by exhibiting $\witness$ and subproofs for
\begin{equation}
  \label{eq:redundance-rule}
  \core \cup \derived \cup \set{\neg C}
  \derives
  \restrict{(\core \cup \derived \cup \set{C})}{\witness}
  \cup
  \set{
    \obj \geq \restrict{\obj}{\witness}
  }
  \eqcomma
\end{equation}
using the previously discussed rules
(where the notation
$\core_1 \derives \core_2$ means that the constraints~$\core_2$ can be derived from
the constraints~$\core_1$).

During preprocessing, constraints in the input formula are often deleted or
replaced by other constraints, in which case the proof should
establish that these deletions maintain equioptimality.
Removing constraints from the derived set~$\derived$ is unproblematic,
but arbitrary deletions from the core set~$\core$ could potentially
introduce
spurious better solutions.
Therefore, removing~$C$   
from~$\core$ can only be done by
the \emph{checked deletion rule}, which requires a proof that the
redundance rule can be used to rederive~$C$ from
$\core \setminus \set{C}$ (see~\cite{BGMN23Dominance} for a more
detailed explanation).

Finally, it turns out to be useful to allow replacing~$\objorig$ by a
new objective~$\objmodified$ using an
\emph{objective function update rule},
as long as this does not change the optimal value of the problem.
Formally, updating the objective from
$\objorig$ to~$\objmodified$
requires derivations of the two constraints
$\objorig \geq \objmodified$
and
$\objmodified \geq \objorig$
from the core set~$\core$,
which shows that any satisfying solution to $\core$ has the same
value for both objectives.
More details on this rule can be found in~\cite{HOGN24CertifyingMIPpresolve}.

%
%

\subsection{Maximum Satisfiability}
\label{subsec:prelim-maxsat}
A WCNF instance of (weighted partial) maximum satisfiability
$\wcnfinstance = (\hards, \softs)$ 
is a conjunction of two CNF formulas 
$\hards$ and $\softs$ with hard and soft clauses, respectively,
where soft clauses~$C \in \softs$ have positive weights~$\weight^C$.
A solution $\rho$ to~$\wcnfinstance$ satisfies~$\hards$ and has cost
$\cost(\softs, \rho)$
equal to the sum of weights of all
soft clauses \emph{not satisfied} by $\rho$.
The optimum $\opt(\wcnfinstance)$ of $\wcnfinstance$ is
$\cost(\softs, \rho)$ of an optimal solution~$\rho$ minimizing the
cost, or $\infty$ if no solution exists.

%
State-of-the-art MaxSAT preprocessors such as
\maxpre~\cite{IBJ22ClauseRedundancy,DBLP:conf/sat/KorhonenBSJ17}
take a slightly different
\emph{objective-centric} view~\cite{DBLP:conf/jelia/BergJ19}
of MaxSAT instances
$\instance \equiv (\formula, \obj)$
as consisting of a CNF formula $\formula$
and an objective function $\obj = \sum_j c_j \ell_j + \obc$
to be minimized under assignments~$\rho$ satisfying~$\formula$.
%
A WCNF MaxSAT instance $\wcnfinstance = (\hards, \softs)$ is converted
into objective-centric  form
$\objectivetransform(\wcnfinstance) = (\formula, \obj)$ 
by letting  the clauses 
$\formula = \hards \cup 
\setdescr{ C \lor b_C}{ C \in \softs,\setsize{C} > 1}$
 of $\objectivetransform(\wcnfinstance)$ 
consist of the hard clauses of  $\wcnfinstance$ and the non-unit soft
clauses in $\softs$, each extended with a 
fresh variable $b_C$ that does not appear in any other clause.
The objective
$\obj =
\sum_{(\olnot{\ell}) \in \softs}
\weight^{(\olnot{\ell})} \ell
+
\sum \weight^C b_C$ 
contains literals~$\ell$
for all unit soft clauses~$\olnot{\ell}$ in~$\softs$ as
literals for all new variables~$b_C$,
with coefficients equal to the weights of the corresponding soft clauses.
In other words, each unit soft clause $(\olnot{\ell}) \in \softs$ of
weight $\weight$ 
is transformed into the term $\weight \cdot \ell$ in the objective
function $\obj$, and each non-unit soft clause~$C$ is transformed into
the hard clause $C \lor b_C$ paired with the 
unit soft clause $(\olnot{b}_C)$ with same weight as~$C$.
The following observation summarizes the properties of
$\objectivetransform(\wcnfinstance)$ that are central to our work.

\begin{observation}\label{prop:pbtransform}
Assume $\rho$ is a solution to a WCNF MaxSAT instance
$\wcnfinstance$. There exists a solution $\rho'$ to  
$(\formula, \obj) = \objectivetransform(\wcnfinstance)$ with
$\cost(\wcnfinstance, \rho) = \obj(\rho')$.  
Conversely, if $\rho_2'$ is a solution to
$\objectivetransform(\wcnfinstance)$, then there exists a solution
$\rho_2$ of $\wcnfinstance$ 
for which $\cost(\wcnfinstance, \rho_2) \leq \obj(\rho_2')$.  
\end{observation}

%
Since objective-centric MaxSAT can be seen to be a pseudo-Boolean
optimization problem, 
Observation~\ref{prop:pbtransform} implies that the optimal value
of the objective~$\obj$ subject to constraints
$\asPB(\formula)$, where
$(\formula, \obj) = \objectivetransform(\wcnfinstance)$, is equal
to the optimal cost of $\wcnfinstance$ (cf.~\cref{fig:enc-correct}
for the formalized theorems).

\section{Proof Logging for MaxSAT Preprocessing}
\label{sec:logging-techniques}

We now discuss how pseudo-Boolean proof logging can be used to reason about correctness of MaxSAT preprocessing steps.
Out approach maintains a correspondence between the working-instance of the preprocessor and the core constraints of the proof.
Specifically, after each preprocessing step (i.e.\ application of a preprocessing technique) the set of derived constraints will be empty.
Thus we focus here on the core set of the proof; adding a constraint to the proof should be understood as adding 
it to the derived set and moving it to the core set, deleting a constraint as using checked deletion to remove it from the core set.
\iflongversion
Full technical details are in Appendix~\ref{apx:prepro}.
\else
Full details are in an online appendix \TODO{cite?}.
\fi

\subsection{Overview}
\begin{figure}[t]
\centering
\begin{tikzpicture}


\node[anchor=west,  align=left] (L1)  at (3,-1)  {\emph{preprocessing} \\ \emph{(MaxSAT)}};
\node[right=2 of L1.east, anchor=west,  align=left] (L2)   {\emph{proof} \\ \emph{(pseudo-Boolean)}};

\node[anchor=west, align=left] (S1) at (0, -2) {\textbf{1. Initialization}};
\node[align=left, anchor=west] at (L1.west |- S1) () {$(\wcnfinstance, 0)$};
\node[align=left, anchor=west] at (L2.west |- S1) (P1) {$(\asPB(\formula^0), \obj^0)$ \\  {\scriptsize where $(\formula^0, \obj^0) = \objectivetransform(\wcnfinstance)$}};

\node[below=1 of S1.west, anchor=west, align=left] (S2) {\textbf{2. Preprocessing} \\ \textbf{on WCNF}};
\node[align=left, anchor=west] at (L1.west |- S2) () {$(\wcnfinstance_1, \textsc{lb}^1)$};
\node[align=left, anchor=west] at (L2.west |- S2) () {$(\core^1, \obj^1)$};

\node[below=1.2 of S2.west, anchor=west, align=left] (S3) {\textbf{3. Conversion to} \\ \textbf{objective-centric}};
\node[align=left, anchor=west] at (L1.west |- S3) () {$(\formula^2, \obj^2 + \textsc{lb}^1 )$ \\ \scriptsize where \\  \scriptsize$(\formula^2, \obj^2) = \objectivetransform(\wcnfinstance_1)$};
\node[align=left, anchor=west] at (L2.west |- S3) () {$(\asPB(\formula^2),\obj^2 + \textsc{lb}^1)$};

\node[below=1.2 of S3.west, anchor=west, align=left] (S4) {\textbf{4. Preprocessing} \\ \textbf{on objective-}\\\textbf{centric}};
\node[align=left, anchor=west] at (L1.west |- S4) () {$(\formula^3, \obj^3)$};
\node[align=left, anchor=west] at (L2.west |- S4) () {$(\asPB(\formula^3), \obj^3)$};

\node[below=1.2 of S4.west, anchor=west, align=left] (S5) {\textbf{5. Constant} \\ \textbf{removal}};
\node[align=left, anchor=west] at (L1.west |- S5) (P6) {$(\formula^4, \obj^4)$ \\ \scriptsize where $\formula^4 = \formula^3 \land (b^{\obc^3})$ \\ \scriptsize $\obj^4 = \obj^3 - \obc^3 + \obc^3 b^{\obc^3}$ };
\node[align=left, anchor=west] at (L2.west |- S5) (P5) {$(\asPB(\formula^4), \obj^4)$};

\node[below=1.4 of S5.west, anchor=west, align=left] (S6) {\textbf{Output}};
\node[align=left, anchor=west, draw] at (L1.west |- S6) () {Preprocessed \\ WCNF $\wcnfinstance_P = (\formula^4, \softs^P)$};
\node[align=left, anchor=west, draw] at (L2.west |- S6) () {Proof of equioptimality \\ of $\asPB(\formula^0)$ under $\obj^0$ \\and $\asPB(\formula^4)$ under $\obj^4$};

\node[draw, dashed,   fit=(L1)(S1) (P1)(P5)(S5)(P6)] {};

\end{tikzpicture}
\caption{Overview of the five stages of certified MaxSAT preprocessing of a WCNF instance $\wcnfinstance$. The middle column contains the state of the working MaxSAT instance as a WCNF instance and a lower bound on its optimum cost (Stages $1$--$2$), or as an objective-centric instance (Stages $3$--$5$). The right column contains a tuple $(\core, \obj)$ with the set $\core$ of core constraints, and objective $\obj$, respectively, of the proof after each stage. 
} \label{fig:prepro-flow}
\end{figure}

We consider
preprocessing of WCNF MaxSAT instances under \emph{equioptimality}---this refers to the application of
steps of
inference (preprocessing) rules to $\wcnfinstance$,  resulting in
an \emph{equioptimal} instance $\wcnfinstance_P$, i.e., an
instance $\wcnfinstance_P$ such that $\opt(\wcnfinstance)=\opt(\wcnfinstance_P)$.
In \emph{certified} MaxSAT preprocessing,
we also provide a proof justifying the claim $\opt(\wcnfinstance)=\opt(\wcnfinstance_P)$.
We discuss inputs that have solutions. Our approach also covers the---arguably less interesting---case
of $\wcnfinstance$ not having solutions;
\iflongversion
details are in Appendix~\ref{apx:nosols}.
\else
details can be found in the online appendix.
\fi



An  overview of the workflow of our certifying
MaxSAT preprocessor is shown in~\cref{fig:prepro-flow}.
Given a WCNF instance $\wcnfinstance$ as input,
the preprocessor proceeds in $5$ stages (illustrated on the left in
 Figure~\ref{fig:prepro-flow}), and then outputs a preprocessed MaxSAT
instance $\wcnfinstance_P$ and a pseudo-Boolean proof which shows that
$\opt(\objectivetransform(\wcnfinstance)) = \opt(\objectivetransform(\wcnfinstance_P))$.
For certified MaxSAT preprocessing,
a formally verified checker (\cref{sec:verified}) can be used on the proof log
to verify that:
\begin{inparaenum}[(a)]
\item the initial core constraints in the
proof correspond exactly to the clauses in $\objectivetransform(\wcnfinstance)$,
\item the final core
constraints in the proof correspond exactly to the clauses in $\objectivetransform(\wcnfinstance_P)$, and
\item that each step in the proof is valid.
\end{inparaenum}
Below, we provide more details on the $5$ sequential stages of the preprocessing flow.

\subsubsection*{Stage 1: Initialization.} An input WCNF instance $\wcnfinstance$
is converted to pseudo-Boolean format by converting it to an objective-centric
representation $(\formula^0, \obj^0) = \objectivetransform(\wcnfinstance)$ and then
representing all clauses in $\formula^0$ as pseudo-Boolean constraints.
The initial proof maintained by the preprocessor will have the constraints
$\asPB(\formula^0)$ as core constraints and $\obj^0$ as the objective.
The initial working MaxSAT instance will be $\wcnfinstance$. Additionally,
the preprocessor maintains a lower bound (initialized to $0$) on the optimal
cost of $\wcnfinstance$.

\subsubsection*{Stage 2: Preprocessing on the Initial WCNF Representation.}
During preprocessing on the WCNF representation, a (very limited) set of simplification
techniques are applied on the working formula. At this stage the preprocessor removes
duplicate, tautological, and blocked clauses~\cite{DBLP:conf/tacas/JarvisaloBH10}.
Additionally, hard unit clauses are unit propagated and clauses subsumed by hard clauses are removed.
Importantly, the preprocessor is performing these simplifications on a WCNF MaxSAT instance where it deals with hard and soft clauses.
As the pseudo-Boolean proof has no concept of hard or soft clauses, the reformulation
steps must be logged in terms of the constraints in the proof.
The next example illustrates how reasoning with different types of clauses is logged in the proof.

\begin{example}
Assume that the working instance of the preprocessor has two duplicate hard clauses $C$ and $D$. Then
the proof has two identical linear inequalities $\asPB(C)$ and $\asPB(D)$ in the core set.
Since $\asPB(D)$ is RUP, it can be deleted directly.
If $D$ is a non-unit soft clause instead, the proof has the linear
constraint $\asPB(D \lor b_D)$ and its objective the term $\weight^D b_D$ where $b_D$ does not appear in any other constraint.
Then we log four steps:
\begin{inparaenum}[(1)]
\item remove the (RUP) constraint $\asPB(D \lor b_D)$,
\item introduce $\olnot{b}_D \geq 1$ to the proof by redundance-based strengthening
using the witness $\{b_D \to 0\}$, 
\item remove the term $\weight^D b_D$ from the objective, and
\item remove the constraint  $\olnot{b}_D \geq 1$ with the witness $\{b_D \to 0\}$.
\end{inparaenum}
\end{example}

\subsubsection*{Stage 3: Conversion to Objective-Centric Representation.}
In order to apply more simplification rules in a cost-preserving way, the
working instance $\wcnfinstance_1 = (\hards^1, \softs^1)$ at the end of Stage 2
is converted into the corresponding objective-centric representation that takes the
lower-bound $\textsc{lb}$ incurred during Stage 1 into account. More specifically,
the preprocessor next converts its working MaxSAT instance into the objective-centric
instance $\instance_2 = (\formula^2, \obj^2 + \textsc{lb})$ where
$(\formula^2, \obj^2) = \objectivetransform(\wcnfinstance_1)$.

Here, it is important to note that at the end of Stage 2, the core constraints
$\core^1$ and objective $\obj^1$ of the proof
are not necessarily $\asPB(\formula^2)$ and $\obj^2 + \textsc{lb}$, respectively.
Specifically,
consider a unit soft clause $(\olnot{\ell})$ of  $\wcnfinstance_1$ obtained
by shrinking  a non-unit soft clause $C \supseteq (\olnot{\ell})$ of the input instance, with weight $\weight^C$.
Then the objective function $\obj^2$ in the preprocessor will include the term $\weight^{C} \ell$
that does not appear in the objective function $\obj^1$ in the proof.
Instead, $\obj^1$ contains the term  $\weight^{C} b_{C}$
and $\core^1$ the constraint $\olnot{\ell} + b_{C} \geq 1$ where $b_{C}$ is the
fresh variable added to $C$ in Stage 1. 
In order to ``sync up'' the  working instance and the proof we log the following four steps:
\begin{inparaenum}[(1)]
\item introduce $\ell + \olnot{b}_C \geq 1$ to the proof with the witness $\{b_C \to 0\}$,
\item update $\obj^1$ by adding $\weight^C \ell- \weight^C b_C$,
\item remove the constraint $\ell + \olnot{b}_C \geq 1$ with the witness $\{b_C \rightarrow 0\}$,
\item remove the constraint $\olnot{\ell} +b_C \ge 1$ with witness $\{b_C \rightarrow 1\}$.
\end{inparaenum}
 The same steps are logged for  all soft unit clauses of $\wcnfinstance_1$ obtained during Stage 2.
 In the last stages, the preprocessor will operate on an objective-centric MaxSAT instance whose clauses correspond exactly to the core constraints of the proof.

\subsubsection*{Stage 4:  Preprocessing on the Objective-Centric Representation.}

During preprocessing on the objective-centric representation, more simplification techniques are applied to the working objective-centric instance and logged to the proof.
We implemented proof logging for a wide range of preprocessing techniques. These include MaxSAT
versions
of rules commonly used
in SAT solving like: bounded variable elimination (BVE)~\cite{EB05EffectivePreprocessing,DBLP:conf/sat/SubbarayanP04a}, bounded variable addition~\cite{MHB12AutomatedReencoding}, blocked clause elimination~\cite{DBLP:conf/tacas/JarvisaloBH10}, subsumption elimination,
self-subsuming resolution~\cite{EB05EffectivePreprocessing,DBLP:conf/cp/OstrowskiGMS02}, failed literal elimination~\cite{Freeman1995ImprovementsPropositionalSatisfiability,DBLP:journals/endm/Berre01,DBLP:conf/aaai/ZabihM88}, and
equivalent literal substitution~\cite{DBLP:journals/tsmc/Brafman04,DBLP:conf/aaai/Li00,DBLP:journals/amai/Gelder05}. We also
cover MaxSAT-specific preprocessing rules like TrimMaxSAT~\cite{DBLP:conf/vmcai/PaxianRB21}, (group)-subsumed
literal (or label) elimination (SLE)~\cite{DBLP:conf/ecai/BergSJ16,DBLP:conf/sat/KorhonenBSJ17},
intrinsic at-most-ones~\cite{DBLP:journals/jsat/IgnatievMM19,IBJ22ClauseRedundancy},
binary core removal (BCR)~\cite{DBLP:conf/focs/Gimpel64,DBLP:conf/sat/KorhonenBSJ17}, label matching~\cite{DBLP:conf/sat/KorhonenBSJ17},
and hardening~\cite{DBLP:conf/cp/AnsoteguiBGL12,IBJ22ClauseRedundancy,DBLP:conf/sat/MorgadoHM12}.
Here we give examples for BVE, SLE, label matching, and BCR, the rest are detailed in
\iflongversion
Appendix~\ref{apx:prepro}.
\else
the online appendix~\cite{}.
\fi
In the following descriptions, let $(\formula, \obj)$ be the current objective-centric working instance.

\paragraph{Bounded Variable Elimination (BVE)~\cite{EB05EffectivePreprocessing,DBLP:conf/sat/SubbarayanP04a}.}
BVE eliminates a variable $x$ that does not appear in the objective from $\formula$ by replacing all clauses in which either $x$ or $\olnot{x}$ appears with the non-tautological clauses in
$\{C \lor D\mid C \lor x \in \formula, D \lor \olnot{x} \in \formula\}$.

An application of BVE is logged as follows:
\begin{inparaenum}[(1)]
\item each non-tautological constraint $\asPB(C\lor D)$ is added by summing the existing constraints $\asPB(C \lor x)$ and $\asPB(D \lor \olnot{x})$ and
\item each constraint of form
$\asPB(C \lor x)$ and $\asPB(D \lor \olnot{x})$ is deleted with the witnesses $x \to 1$ and $x \to 0$, respectively.
\end{inparaenum}


\paragraph{Label Matching~\cite{DBLP:conf/sat/KorhonenBSJ17}.}
Label matching allows merging pairs of objective variables that can be deduced to not both be set to $1$ by optimal solutions.
Assume that
\begin{inparaenum}[(i)]
\item $\formula$ contains the clauses $C\lor b_C$ and $D\lor b_D$,
\item $b_C$ and $b_D$ are objective variables with the same coefficient $\weight$ in $\obj$, and
\item $C\lor D$ is a tautology.
\end{inparaenum}
Then label matching replaces $b_C$ and $b_D$ with a fresh variable $b_{CD}$, i.e., replaces $C\lor b_C$ and $D\lor b_D$ with $C\lor b_{CD}$ and $D\lor b_{CD}$ and adds $-wb_C-wb_D+wb_{CD}$ to $\obj$.

As $C \lor D$ is a tautology we assume w.l.o.g.~that $\bar{x}\in C$ and  $x \in D$ for a variable $x$.
Label matching is logged via the following steps:
\begin{inparaenum}[(1)]
\item introduce the constraint $\olnot{b}_C + \olnot{b}_D \geq 1$ with the witness $\{b_C\rightarrow x, b_D\rightarrow \olnot{x}\}$,
\item introduce the constraints $b_{CD} + \olnot{b}_C + \olnot{b}_D \geq 2$ and $\olnot{b}_{CD} + b_C + b_D \geq 1$ by reification; these correspond to
$b_{CD} = b_C + b_D$ which holds even though the variables are binary due to the constraint added in the first step,
\item update the objective by adding $-wb_C-wb_D+wb_{CD}$ to it,
\item introduce the constraints $\asPB(C\lor b_{CD})$ and $\asPB(D\lor b_{CD})$ which are RUP,
\item delete $\asPB(C\lor b_C)$ and $\asPB(D\lor b_D)$ with the witness $\{b_C\rightarrow \olnot{x}, b_D\rightarrow x\}$,
\item delete the constraint $b_{CD} + \olnot{b}_C + \olnot{b}_D \geq 2$ with the witness $\{b_C \to 0, b_D \to 0\}$  and $\olnot{b}_{CD} + b_C + b_D \geq 1$ with the witness $\{b_C \to 1, b_D \to 0\}$,
\item delete $\olnot{b}_C + \olnot{b}_D \geq 1$ with the witness $\{b_C \to 0\}$.
\end{inparaenum}

\paragraph{Subsumed Literal Elimination (SLE)~\cite{DBLP:conf/ecai/BergSJ16,IBJ22ClauseRedundancy}.}
Given two non-objective variables $x$ and $y$ s.t.
\begin{inparaenum}[(i)]
\item \label{itm:sle1} $\{ C \mid  C \in \formula, y \in C\} \subseteq \{ C \mid  C \in \formula, x \in C\}$ and
\item \label{itm:sle2} $\{ C \mid  C \in \formula, \olnot{x} \in C\} \subseteq \{ C \mid  C \in \formula, \olnot{y} \in C\}$
\end{inparaenum},
subsumed literal elimination (SLE) allows fixing $x=1$ and $y=0$.
SLE is logged with the following steps:
\begin{inparaenum}[(1)]
\item introduce $x\ge 1$ and $\olnot{y}\ge 1$, both with  witness $\{x\rightarrow 1$,  $y\rightarrow 0\}$,
\item simplify the constraint database via propagation,
\item delete the constraints introduced in the first step as neither $x$ nor $y$ appears in any other constraints after the simplification.
\end{inparaenum}

If $x$ and $y$ are objective variables, the application of SLE additionally requires that:
\begin{inparaenum}[(i)]
\setcounter{enumi}{2}
\item \label{itm:sle3} the coefficient in the objective of $x$ is at most as high as the coefficient of $y$.
\end{inparaenum}
Then
the value of $x$ is not fixed as it would incur cost. Instead, only $y=0$ is fixed and $y$ removed from the objective.
Intuitively, conditions (\ref{itm:sle1}) and (\ref{itm:sle2}) establish that the values of $x$ and $y$ can always be flipped to $0$ and $1$, respectively,
without falsifying any clauses. If neither of the variables is in the objective, this flip does not increase the
cost of any solutions. Otherwise, condition (\ref{itm:sle3}) ensures that the flip does not make the solution worse, i.e., increase its cost.

\paragraph{Binary Core Removal (BCR)~\cite{DBLP:conf/focs/Gimpel64,DBLP:conf/sat/KorhonenBSJ17}.}
Assume that the following four prerequisites hold:
\begin{inparaenum}[(i)]
\item $\formula$ contains a clause $b_C \lor b_D$ for two objective variables $b_C$ and $b_D$,
\item $b_C$ and $b_D$ have the same coefficient $w$ in $\obj$,
\item the negations $b_C$ and $b_D$ do not appear in any clause of $F$, and
\item both $b_C$ and $b_D$ appear in at least one other clause of $F$ but not together in any other clause of $F$.
\end{inparaenum}
Binary core removal replaces all clauses containing $b_C$ or $b_D$ with the non-tautological clauses in $\{ C \lor  D \lor b_{CD} \mid C \lor b_C\in \formula, D \lor b_D\in \formula\}$, where $b_{CD}$ is a fresh variable, and modifies the objective function by adding $-w b_C -w b_D +wb_{CD}+w$ to it.

BCR is logged as a combination of the so-called \emph{intrinsic at-most-ones} technique~\cite{DBLP:journals/jsat/IgnatievMM19,IBJ22ClauseRedundancy} and BVE.
Applying intrinsic at most ones on the variables $b_C$ and $b_D$ introduces a new clause $(\olnot{b}_C \lor \olnot{b}_D  \lor b_{CD})$ and adds
$-w b_C - w b_D + w b_{CD} + w$ to the objective. Our proof for intrinsic at most ones is the same as the one presented in~\cite{BBNOV23CertifiedCoreGuided}.
As this step removes $b_C$ and $b_D$ from the objective, both can now be eliminated via BVE.

\subsubsection*{Stage 5: Constant Removal and Output.}\label{sec:mprepro5}

After objective-centric preprocessing, the
final objective-centric instance $(\formula^3, \obj^3)$ is converted back to
a WCNF instance. Before doing so, the constant term $\obc_3$ of $\obj^3$ is removed by introducing a fresh variable $b^{\obc_3}$,
and setting $\formula^4 = \formula^3 \land (b^{\obc_3})$ and $\obj^4 = \obj^3 - \obc_3 + \obc_3 b^{\obc_3}$. This
step is straightforward to prove.


Finally, the preprocessor outputs the WCNF instance $\wcnfinstance_P = (\formula^4, \softs^P)$ that
has $\formula^4$ as hard clauses.
the set $\softs^P$ of soft clauses consists of a unit soft clause $(\olnot{\ell})$ of weight
$c$ for each term $c \cdot \ell$ in $\obj^4$. The preprocessor also outputs the final proof of
the fact  that the minimum-cost of solutions to the
pseudo-Boolean formula $\asPB(\formula^0)$ under $\obj^0$ is the
same as that of $\asPB(\formula^4)$ under $\obj^4$, i.e. that
$\opt(\objectivetransform(\wcnfinstance)) = \opt(\objectivetransform(\wcnfinstance_P))$.



\newcommand{\tablesubsectionA}[1]{ \iffalse \hline \multicolumn{5}{l}{\textbf{#1}} \\ \hline \fi }
\newcommand{\tablesubsectionB}[1]{ \hline & \multicolumn{4}{l}{\textit{#1}} \\ \hline}

\subsection{Worked Example of Certified Preprocessing}
\begin{table}[t]
\caption{Example proof produced by a certifying preprocessor. The column (ID) refers
to constraint IDs in the pseudo-Boolean proof. The column (Step) indexes all proof logging steps and is used
when referring to the steps in the discussion. The letter $\omega$ is used for the witness substitution
in redundance-based strengthening steps.
	\label{tbl:exampleproof}}
\centering
\setlength\tabcolsep{9pt}
\begin{tabular}{ c | c | r | l | r}

\textbf{Step} &  \textbf{ID} &  \textbf{Type} & \textbf{Justification} & \textbf{Objective} \\
 \hline
1 & (1) 	& add $x_1 + x_2 \geq 1$ & input &  $x_1 + 2b_1 + 3b_2$  \\
2 & (2) 	& add $\olnot{x}_2 \geq 1$ & input &  $x_1 + 2b_1 + 3b_2$ \\
3 & (3) 	& add $x_3 + \olnot{x}_4 + b_1 \geq 1$ & input &  $x_1 + 2b_1 + 3b_2$ \\
4 & (4) 	& add $x_4 + \olnot{x}_5 + b_2 \geq 1$ & input &  $x_1 + 2b_1 + 3b_2$ \\
\tablesubsectionB{Unit propagation: fix $x_2=0$, constraint (2)}
5 & (5) 	& add $x_1 \geq 1$ & $(1) + (2)$ & $x_1 + 2b_1 + 3b_2$ \\
6 &       	& delete (1) & RUP & $x_1 + 2b_1 + 3b_2$ \\
7 &      	& delete (2) & $\omega \colon \{x_2 \to 0\}$ & $x_1 + 2b_1 + 3b_2$ \\
\tablesubsectionB{Unit propagation; fix $x_1=1$, constraint (5)}
8 &      	& add $-x_1 + 1$ to obj. & $(5)$  & $2b_1 + 3b_2 + 1$\\
9 &      	& delete (5) &  $\omega \colon \{x_1 \to 1\}$ & $2b_1 + 3b_2 + 1$\\
\tablesubsectionB{BVE: eliminate $x_4$}
10 & (6) 	& \makecell[l]{ add \\$x_3 + b_1 + \olnot{x}_5 + b_2 \geq 1$} & $(3) + (4)$  & $2b_1 + 3b_2 + 1$ \\
11 &	& delete (3) & $\omega \colon \{x_4 \to 0\}$ &  $2b_1 + 3b_2 + 1$ \\
12 &	& delete (4) & $\omega \colon \{x_4 \to 1\}$ &  $2b_1 + 3b_2 + 1$ \\
\tablesubsectionB{Subsumed literal elimination: $\olnot{b}_2$}
13 & (7) 	& add $\olnot{b}_2 \geq 1$ & $\omega \colon \{b_2 \to 0, b_1 \to 1\}$ & $2b_1 + 3b_2 + 1$ \\
14 & 	& add $-3b_2$ to obj. & $(7)$ & $2b_1 + 1$\\
15 & (8) &  add $x_3 + b_1 + \olnot{x}_5 \geq 1$ & $(6) + (7)$ & $2b_1 + 1$ \\
16 & & delete (6) & RUP & $2b_1 + 1$ \\
17 & 	& delete (7) & $\omega \colon \{b_2 \to 0\}$  & $2b_1 + 1$\\
\tablesubsectionB{Remove objective constant}
18 & (9) 	&add $b_3 \geq 1$ 	&  $\omega \colon \{b_3 \to 1\}$ & $2b_1 + 1$ \\
19 & 	& add $b_3 - 1$ to obj. & $(9) $ & $2b_1 + b_3$
\end{tabular}
\end{table}

We give a worked-out example of certified preprocessing
of the instance $\wcnfinstance = (\hards, \softs)$
where $\hards = \{ (x_1 \lor x_2), (\olnot{x}_2)\}$ and
three soft clauses: $(\olnot{x}_1)$ with weight $1$,
$(x_3 \lor  \olnot{x}_4)$ with weight $2$, and $ (x_4 \lor \olnot{x}_5)$ with weight $3$.
The proof for one possible execution of the preprocessor on this input instance is detailed in Table~\ref{tbl:exampleproof}.

During Stage 1 (Steps 1--4 in Table~\ref{tbl:exampleproof}), the core constraints of the proof are initialized to contain the four constraints
corresponding to the hard and non-unit soft clauses
of $\wcnfinstance$ (IDs (1)--(4) in Table~\ref{tbl:exampleproof}),
and the objective to $x_1 + 2b_1 + 3b_2$, where $b_1$ and $b_2$ are fresh variables
added to the non-unit soft clauses of $\wcnfinstance$.

During Stage 2 (Steps 5--9), the preprocessor fixes $x_2 = 0$
via unit propagation by removing $x_2$ from the clause $(x_1 \lor x_2)$, and then
removing the unit clause $(\olnot{x}_2)$. The justification for fixing $x_2 = 0$ are Steps 5--7.
Next the preprocessor fixes $x_1=1$ which (i) removes the hard clause $(x_1)$, and (ii) increases the lower bound
on the optimal cost by $1$. The justification for fixing $x_1 = 1$ are Steps 8 and 9 of Table~\ref{tbl:exampleproof}.
At this point---at the end of Stage 2---the working instance $\wcnfinstance_1 = (\hards^1, \softs^1)$
has $\hards^1 = \{\}$ and $\softs^1 = \{ (x_3 \lor  \olnot{x}_4), (x_4 \lor \olnot{x}_5)\}$.

In Stage 3, the preprocessor converts its working instance into the objective-centric representation $(\formula, \obj)$
where $\formula = \{  (x_3 \lor  \olnot{x}_4 \lor b_1), (x_4 \lor \olnot{x}_5 \lor b_2)\}$ and $\obj = 2b_1 + 3b_2 + 1$, which
exactly matches the core constraints and objective of the proof after Step 9.
Thus, in this instance, the conversion does not result in any proof logging steps.
Afterwards, during Stage 4 (Steps 10--17),
the preprocessor applies BVE in order to eliminate $x_4$ (Steps 10--12) and SLE to fix
$b_2$ to $0$ (Steps 13--17). Finally,
Steps 18 and 19 represent Stage 5, i.e.,\  the removal of the constant $1$ from the objective.
After these steps, the preprocessor outputs the preprocessed instance $\wcnfinstance_P = (\hards^P, \softs^P)$,
where $\hards^P = \{ (x_3 \lor \olnot{x}_5 \lor b_1), (b_3)\}$ and $\softs^P$ contains two clauses: $(\olnot{b}_1)$
with weight $2$, and $(\olnot{b}_3)$ with weight $1$.

\section{Verified Proof Checking for Preprocessing Proofs}
\label{sec:verified}

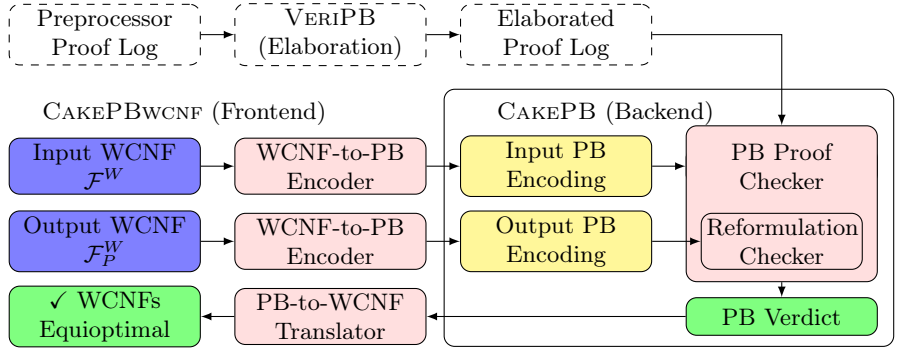
\begin{figure*}[t]
  \centering
  \begin{tikzpicture}[
    node distance=0.5cm and 3cm,
    every node/.style={font=\footnotesize},
    input/.style={fill=blue!50, draw},
    process/.style={fill=pink!50, draw},
    external/.style={dashed},
    file/.style={fill=yellow!50, draw},
    result/.style={fill=green!50, draw}]

    \coordinate (row0);
    \coordinate [above=of row0] (rown1);
    \coordinate [below=of row0] (row1);
    \coordinate [below=of row1] (row2);
    \coordinate [below=of row2] (row3);
    \coordinate [below=of row3] (row4);
    \coordinate [below=of row4] (row5);
    \coordinate [below=of row5] (row6);
    \coordinate [below=of row6] (row7);
    \coordinate (rown05) at ($(rown1)!0.5!(row0)$);
    \coordinate (row05) at ($(row0)!0.5!(row1)$);
    \coordinate (row15) at ($(row1)!0.5!(row2)$);
    \coordinate (row25) at ($(row2)!0.5!(row3)$);
    \coordinate (row35) at ($(row3)!0.5!(row4)$);
    \coordinate (row45) at ($(row4)!0.5!(row5)$);
    \coordinate (row55) at ($(row5)!0.5!(row6)$);
    \coordinate (row65) at ($(row6)!0.5!(row7)$);
    \coordinate (col1) at (row1);
    \coordinate [right=of col1] (col2);
    \coordinate [right=of col2] (col3);
    \coordinate [right=of col3] (col4);
    \coordinate (col15) at ($(col1)!0.5!(col2)$);
    \coordinate (col25) at ($(col2)!0.5!(col3)$);
    \coordinate (col275) at ($(col25)!0.5!(col3)$);
    \coordinate (col35) at ($(col3)!0.5!(col4)$);
    \coordinate (col375) at ($(col35)!0.5!(col4)$);

    \begin{scope}[every node/.style={rectangle, rounded corners, text centered, minimum height=0.5cm,
        inner sep=2pt, text width=2.4cm, text centered, on grid, font=\footnotesize}]
      \node [input] at (col1|-row3) (wcnfinput) {Input WCNF\\$\wcnfinstance$};
      \node [external,draw,dashed] (augmentedproof)  at (col1|-rown05) {Preprocessor \\ Proof Log};
      \node [external,draw,dashed] (kernelproof)  at (col3|-rown05) {Elaborated Proof Log};
      \node [external,draw,dashed] (veripb)  at (col2|-rown05) {\veripb (Elaboration)};
      \node [input] at (col1|-row5) (wcnfoutput) {Output WCNF\\$\wcnfinstance_P$};

      \node [process] at (col2|-row3) (encwcnf) {WCNF-to-PB\\Encoder};
      \node [process] at (col2|-row5) (encwcnf2) {WCNF-to-PB\\Encoder};

      \node [result] at (col1|-row7) (wcnfconcl) {{\checkmark} WCNFs Equioptimal};
      \node [process] at (col2|-row7) (translator) {PB-to-WCNF\\Translator};

      \node [file] (pbencoding) at (col3|-row3) {Input PB Encoding};
      \node [result] (pbconcl)  at (col4|-row7) {PB Verdict};

      \node [process] (pbcheckdummy)  at (col4|-row3) {};
      \node [process] (pbcheck)  at (col4|-row4) {~\\PB Proof\\Checker\\~\\~\\~\\};
      \node [process,text width=2cm] (pbcheck2)  at (col4|-row5) {Reformulation\\Checker};

      \node [file] (pbencoding2) at (col3|-row5) {Output PB Encoding};

      \node (cakepbtext) [above=0cm of pbencoding.center, anchor=west, text width=2.9cm,
      minimum height=10pt, inner sep=0pt, yshift=7mm, xshift=-8mm] {\cakepb (Backend)};

      \node (cakepbgraphtext) [above=0cm of wcnfinput.center, anchor=west, text width=4.1cm,
      minimum height=10pt, inner sep=0pt, yshift=7mm, xshift=-10mm] {\cakepbwcnf (Frontend)};

      \begin{scope}[on background layer]
        \node [draw, rounded corners,fit=(pbconcl) (pbencoding) (pbcheck) (cakepbtext), inner xsep=6pt, inner ysep=4pt] (contrib1) {};
      \end{scope}

    \end{scope}

    \draw [-latex] (encwcnf.east)  --++(0:3mm)|- (pbencoding.west);
    \draw [-latex] (encwcnf2.east)  --++(0:3mm)|- (pbencoding2.west);

    \draw [-latex] (wcnfinput.east) --++(0:3mm)|- (encwcnf.west);
    \draw [-latex] (wcnfoutput.east) --++(0:3mm)|- (encwcnf2.west);

    \draw [-latex] (pbencoding.east) to  (pbcheckdummy.west);
    \draw [-latex] (pbcheck.south) to (pbconcl.north);
    \draw [-latex] (pbconcl.west) to (translator.east);
    \draw [-latex] (translator.west) to (wcnfconcl.east);
    \draw [-latex] (pbencoding2.east) to  (pbcheck2.west);

    \draw [-latex] (augmentedproof.east) to  (veripb.west);
    \draw [-latex] (veripb.east) to  (kernelproof.west);
    \draw [-latex] (kernelproof.east) -| (pbcheck.north);

  \end{tikzpicture}
  \caption{Workflow for end-to-end verified MaxSAT preprocessing proof checking.}
  \label{figure:verifiedworkflow}
\end{figure*}

This section presents our new workflow for formally verified, end-to-end proof checking of MaxSAT preprocessing proofs based on pseudo-Boolean reasoning; an overview of this workflow is shown in~\cref{figure:verifiedworkflow}.
To realize this workflow, we extended the \veripb tool and its proof format to support a new \emph{output section} for declaring (and checking) reformulation guarantees between input and output PBO instances (\cref{subsec:output-sec});
we similarly modified \cakepb~\cite{GMMNOT24Subgraph} a verified proof checker to support the updated proof format (\cref{subsec:verified-output});
finally, we built a verified frontend, \cakepbwcnf, which mediates between MaxSAT WCNF instances and PBO instances (\cref{subsec:verified-frontend}).
Our formalization is carried out in the \HOL proof assistant~\cite{SL08BriefOverviewHOL4} using \CakeML tools~\cite{GMKN2017Verified,MO14Proof-Producing,TMKFON19CakeML} to obtain a verified executable implementation of \cakepbwcnf. A snapshot of the \cakepb development is at \url{https://github.com/CakeML/cakeml/tree/9c063a7d92cadb36c49a6e45856ae0db734dc9e5/examples/pseudo_bool}.

In the workflow (\cref{figure:verifiedworkflow}), a MaxSAT preprocessor produces a reformulated output WCNF along with a proof log claiming that it is equioptimal to the input WCNF.
The proof log is elaborated by \veripb then checked by \cakepbwcnf, resulting in a verified \emph{verdict}---if the proof checking succeeds, then the input and output WCNFs are equioptimal.
This workflow also supports verified checking of WCNF MaxSAT solving proofs (where the output parts of the flow are omitted).

\subsection{Output Section for Pseudo-Boolean Proofs}
\label{subsec:output-sec}

Given an input PBO instance $(\formula,\obj)$, the \veripb proof system as described in~\cref{subsec:prelim-veripb} maintains the invariant that the core constraints~$\core$ (and current objective) are equioptimal to the input instance.
Utilizing this invariant, the new \emph{output section} for \veripb proofs allows users to optionally
specify
an output PBO instance $(\formula',\obj')$ at the end of a proof.
This output instance is claimed to be a reformulation of the input which is either:
\begin{inparaenum}[(i)]
\item \emph{derivable}, i.e., satisfiability of $\formula$ implies satisfiability of $\formula'$,
\item \emph{equisatisfiable} to $\formula$, or
\item \emph{equioptimal} to $(\formula,\obj)$.
\end{inparaenum}
These are increasingly stronger claims about the relationship between the input and output instances.
After checking a pseudo-Boolean derivation, \veripb runs reformulation checking which, e.g., for equioptimality, checks that $\core \subseteq \formula'$, $\formula' \subseteq \core$, and that the respective objective functions are syntactically equal after normalization; other reformulation guarantees are checked analogously.

The \veripb tool supports an \emph{elaboration} mode~\cite{GMMNOT24Subgraph}, where it converts a proof in \emph{augmented} proof format to a \emph{kernel} proof format.
The
augmented
format contains syntactic sugar rules to facilitate proof logging for solvers and preprocessors like \maxpre,
while the
kernel
format is supported by the formally verified proof checker \cakepb.
The new output section is passed unchanged from augmented to kernel format during elaboration.

\subsection{Verified Proof Checking for Reformulations}
\label{subsec:verified-output}

There are two main verification tasks involved in extending \cakepb with support for the output section.
The first task is to verify soundness of all cases of reformulation checking.
Formally, the equioptimality of an input PBO instance \HOLFreeVar{fml}, \HOLFreeVar{obj} and its output counterpart \HOLFreeVar{fml'}, \HOLFreeVar{obj'} is specified as follows:
\begin{equation*}
\begin{holthmenv}
\HOLConst{sem_output}\;\HOLFreeVar{fml}\;\HOLFreeVar{obj}\;\HOLConst{None}\;\ensuremath{\HOLFreeVar{fml}\sp{\prime{}}}\;\ensuremath{\HOLFreeVar{obj}\sp{\prime{}}}\;\HOLConst{Equioptimal}\;\HOLTokenDefEquality{}\\
\;\;\HOLSymConst{\HOLTokenForall{}}\HOLBoundVar{v}.\;(\HOLSymConst{\HOLTokenExists{}}\HOLBoundVar{w}.\;\HOLConst{satisfies}\;\HOLBoundVar{w}\;\HOLFreeVar{fml}\;\HOLSymConst{\HOLTokenConj{}}\;\HOLConst{eval_obj}\;\HOLFreeVar{obj}\;\HOLBoundVar{w}\;\HOLSymConst{\HOLTokenLeq{}}\;\HOLBoundVar{v})\;\HOLSymConst{\HOLTokenEquiv{}}\\
\;\;\;\;\;\;\;\;\,(\HOLSymConst{\HOLTokenExists{}}\ensuremath{\HOLBoundVar{w}\sp{\prime{}}}.\;\HOLConst{satisfies}\;\ensuremath{\HOLBoundVar{w}\sp{\prime{}}}\;\ensuremath{\HOLFreeVar{fml}\sp{\prime{}}}\;\HOLSymConst{\HOLTokenConj{}}\;\HOLConst{eval_obj}\;\ensuremath{\HOLFreeVar{obj}\sp{\prime{}}}\;\ensuremath{\HOLBoundVar{w}\sp{\prime{}}}\;\HOLSymConst{\HOLTokenLeq{}}\;\HOLBoundVar{v})
\end{holthmenv}
\end{equation*}

This definition says that, for all values \HOLFreeVar{v}, the input instance has a satisfying assignment with objective value less than or equal to \HOLFreeVar{v} iff the output instance also has such an assignment; note that this implies (as a special case) that \HOLFreeVar{fml} is satisfiable iff \HOLFreeVar{fml'} is satisfiable.
The verified correctness theorem for {\cakepb} says that \emph{if} {\cakepb} successfully checks a pseudo-Boolean proof in kernel format and prints a verdict declaring equioptimality, then the input and output instances are indeed equioptimal as defined in \HOLConst{sem_output}.

The second task is to develop verified optimizations to speedup proof steps which occur frequently in preprocessing proofs; some code hotspots were also identified by profiling the proof checker against proofs generated by {\maxpre}.
Similar (unverified) versions of these optimizations are also used in \veripb.
These optimizations turned out to be necessary in practice---they mostly target steps which, when na\"ively implemented, have quadratic (or worse) time complexity in the size of the constraint database.


\paragraph{Optimizing Reformulation Checking.} The most expensive step in reformulation checking for the output section is to ensure that the core constraints $\core$ are included in the output formula and vice versa (possibly with permutations and duplicity).
Here, {\cakepb} normalizes all pseudo-Boolean constraints involved to a canonical form and then copies both $\core$ and the output formula into respective array-backed hash tables for fast membership tests.

\paragraph{Optimizing Redundance and Checked Deletion Rules.}
A na\"ive implementation of these two redundance-based rules would require walking over the entire database of active constraints, e.g., checking all the subproofs in~\cref{eq:redundance-rule} by looping over all constraints in the RHS $\restrict{(\core \cup \derived \cup \set{C})}{\omega} \cup \set{ \obj \geq \restrict{\obj}{\omega} }$.
An important observation here is that preprocessing proofs frequently use substitutions $\omega$ that only involve a small number of variables (usually one and the variable is fresh for reifications).
Consequently, most of the constraints $\restrict{(\core \cup \derived \cup \set{C})}{\omega}$ can be skipped when checking redundancy because they are unchanged by the substitution.
Similarly, the constraint $\set{\obj \geq \restrict{\obj}{\omega}}$ is expensive to construct when $\obj$ contains many terms, but this construction can be skipped if it is \emph{a priori} known that no variables being substituted occur in $\obj$.
To handle these cases, {\cakepb} stores a lazily-updated mapping of variables to their occurrences in the constraint database and the objective, which it uses to detect and skip the aforementioned cases.

The aforementioned occurrence mapping is necessary for performance due to the
frequency of steps involving witnesses for preprocessing proofs.  However,
storing this mapping does lead to a memory overhead in our checker.  More
precisely, every variable \emph{occurrence} in any constraint in the database
corresponds to exactly one ID in the mapping. Thus, the overhead of storing the
mapping is worst-case quadratic in the number of clauses, but it is still
linear in the total space usage for the constraints database.

\subsection{Verified WCNF Frontend}
\label{subsec:verified-frontend}

The \cakepbwcnf frontend mediates between MaxSAT WCNF problems and pseudo-Boolean optimization problems native to \cakepb.
Accordingly, the correctness of \cakepbwcnf is stated in terms of MaxSAT semantics, i.e., the encoding, underlying pseudo-Boolean semantics, and proof system are all formally verified.
In order to trust \cakepbwcnf, one \emph{only} has to carefully inspect the formal definition of MaxSAT semantics shown in~\cref{fig:maxsatsem} to make sure that it matches the informal definition in~\cref{subsec:prelim-maxsat}.
Here, each clause \HOLFreeVar{C} is paired with a natural number \HOLFreeVar{n}, where $\HOLFreeVar{n} = 0$ indicates a hard clause and when $\HOLFreeVar{n} > 0$ it is the weight of \HOLFreeVar{C}.
The optimal cost of a weighted CNF formula \HOLFreeVar{wfml} is \HOLConst{None} (representing $\infty$) if no satisfying assignment to the hard clauses exist; otherwise, it is the minimum cost among all satisfying assignments to the hard clauses.

\begin{figure}[t]
\begin{holthmenv}
\HOLConst{sat_hard}\;\HOLFreeVar{w}\;\HOLFreeVar{wfml}\;\HOLTokenDefEquality{}\;\HOLSymConst{\HOLTokenForall{}}\HOLBoundVar{C}.\;\HOLConst{mem}\;(\HOLNumLit{0}\HOLSymConst{,}\HOLBoundVar{C})\;\HOLFreeVar{wfml}\;\HOLSymConst{\HOLTokenImp{}}\;\HOLConst{sat_clause}\;\HOLFreeVar{w}\;\HOLBoundVar{C}
\end{holthmenv}

\begin{holthmenv}
\HOLConst{weight_clause}\;\HOLFreeVar{w}\;(\HOLFreeVar{n}\HOLSymConst{,}\HOLFreeVar{C})\;\HOLTokenDefEquality{}\;\HOLKeyword{if}\;\HOLConst{sat_clause}\;\HOLFreeVar{w}\;\HOLFreeVar{C}\;\HOLKeyword{then}\;\HOLNumLit{0}\;\HOLKeyword{else}\;\HOLFreeVar{n}
\end{holthmenv}

\begin{holthmenv}
\HOLConst{cost}\;\HOLFreeVar{w}\;\HOLFreeVar{wfml}\;\HOLTokenDefEquality{}\;\HOLConst{sum}\;(\HOLConst{map}\;(\HOLConst{weight_clause}\;\HOLFreeVar{w})\;\HOLFreeVar{wfml})
\end{holthmenv}

\begin{holthmenv}
\HOLConst{opt_cost}\;\HOLFreeVar{wfml}\;\HOLTokenDefEquality{}\;\HOLKeyword{if}\;\HOLSymConst{\HOLTokenNeg{}}\HOLSymConst{\HOLTokenExists{}}\HOLBoundVar{w}.\;\HOLConst{sat_hard}\;\HOLBoundVar{w}\;\HOLFreeVar{wfml}\;\HOLKeyword{then}\;\HOLConst{None}\\
\;\;\;\;\;\;\;\;\;\;\;\;\;\;\;\;\;\;\;\;\;\;\;\;\;\;\;\;\;\HOLKeyword{else}\;\HOLConst{Some}\;(\HOLConst{min\ensuremath{_{\text{set}}}}\;\HOLTokenLeftbrace{}\HOLConst{cost}\;\HOLBoundVar{w}\;\HOLFreeVar{wfml}\;\HOLTokenBar{}\;\HOLConst{sat_hard}\;\HOLBoundVar{w}\;\HOLFreeVar{wfml}\HOLTokenRightbrace{})
\end{holthmenv}
\caption{Formalized semantics for MaxSAT WCNF problems.}
\label{fig:maxsatsem}
\end{figure}

\paragraph{There and Back Again.}
\cakepbwcnf contains a verified WCNF-to-PB encoder implementing the encoding described in~\cref{subsec:prelim-maxsat}.
Its correctness theorems are shown in~\cref{fig:enc-correct}, where the two lemmas in the top row relate the satisfiability and cost of the WCNF to its PB optimization counterpart after running \HOLConst{wcnf_to_pbf} (and vice versa), see Observation~\ref{prop:pbtransform}.
Using these lemmas, the final theorem (bottom row) shows that equioptimality for two (encoded) PB optimization problems can be \emph{translated} back to equioptimality for the input and preprocessed WCNFs.

\begin{figure}[t]
\begin{minipage}[t]{.47\textwidth}
\begin{holthmenv}
\HOLTokenTurnstile{}\;\HOLConst{wfml_to_pbf}\;\HOLFreeVar{wfml}\;\HOLSymConst{=}\;(\HOLFreeVar{obj}\HOLSymConst{,}\HOLFreeVar{pbf})\;\HOLSymConst{\HOLTokenConj{}}\\
\;\;\;\HOLConst{satisfies}\;\HOLFreeVar{w}\;(\HOLConst{set}\;\HOLFreeVar{pbf})\;\HOLSymConst{\HOLTokenImp{}}\\
\;\;\;\;\;\HOLSymConst{\HOLTokenExists{}}\ensuremath{\HOLBoundVar{w}\sp{\prime{}}}.\;\HOLConst{sat_hard}\;\ensuremath{\HOLBoundVar{w}\sp{\prime{}}}\;\HOLFreeVar{wfml}\;\HOLSymConst{\HOLTokenConj{}}\\
\;\;\;\;\;\;\;\;\;\;\;\;\;\HOLSymConst{}\HOLConst{cost}\;\ensuremath{\HOLBoundVar{w}\sp{\prime{}}}\;\HOLFreeVar{wfml}\;\HOLSymConst{\HOLTokenLeq{}}\;\HOLConst{eval_obj}\;\HOLFreeVar{obj}\;\HOLFreeVar{w}
\end{holthmenv}
\end{minipage}\hfill
\begin{minipage}[t]{.47\textwidth}
\begin{holthmenv}
\HOLTokenTurnstile{}\;\HOLConst{wfml_to_pbf}\;\HOLFreeVar{wfml}\;\HOLSymConst{=}\;(\HOLFreeVar{obj}\HOLSymConst{,}\HOLFreeVar{pbf})\;\HOLSymConst{\HOLTokenConj{}}\\
\;\;\;\HOLConst{sat_hard}\;\HOLFreeVar{w}\;\HOLFreeVar{wfml}\;\HOLSymConst{\HOLTokenImp{}}\\
\;\;\;\;\;\HOLSymConst{\HOLTokenExists{}}\ensuremath{\HOLBoundVar{w}\sp{\prime{}}}.\;\HOLConst{satisfies}\;\ensuremath{\HOLBoundVar{w}\sp{\prime{}}}\;(\HOLConst{set}\;\HOLFreeVar{pbf})\;\HOLSymConst{\HOLTokenConj{}}\\
\;\;\;\;\;\;\;\;\;\;\;\;\;\HOLConst{eval_obj}\;\HOLFreeVar{obj}\;\ensuremath{\HOLBoundVar{w}\sp{\prime{}}}\;\HOLSymConst{=}\;\HOLSymConst{}\HOLConst{cost}\;\HOLFreeVar{w}\;\HOLFreeVar{wfml}
\end{holthmenv}
\end{minipage}~\\~\\

\begin{holthmenv}
\HOLTokenTurnstile{}\;\HOLConst{full_encode}\;\HOLFreeVar{wfml}\;\HOLSymConst{=}\;(\HOLFreeVar{obj}\HOLSymConst{,}\HOLFreeVar{pbf})\;\HOLSymConst{\HOLTokenConj{}}\;\HOLConst{full_encode}\;\ensuremath{\HOLFreeVar{wfml}\sp{\prime{}}}\;\HOLSymConst{=}\;(\ensuremath{\HOLFreeVar{obj}\sp{\prime{}}}\HOLSymConst{,}\ensuremath{\HOLFreeVar{pbf}\sp{\prime{}}})\;\HOLSymConst{\HOLTokenConj{}}\\
\;\;\;\HOLConst{sem_output}\;(\HOLConst{set}\;\HOLFreeVar{pbf})\;\HOLFreeVar{obj}\;\HOLConst{None}\;(\HOLConst{set}\;\ensuremath{\HOLFreeVar{pbf}\sp{\prime{}}})\;\ensuremath{\HOLFreeVar{obj}\sp{\prime{}}}\;\HOLConst{Equioptimal}\;\HOLSymConst{\HOLTokenImp{}}\\
\;\;\;\;\;\HOLConst{opt_cost}\;\HOLFreeVar{wfml}\;\HOLSymConst{=}\;\HOLConst{opt_cost}\;\ensuremath{\HOLFreeVar{wfml}\sp{\prime{}}}
\end{holthmenv}
\caption{Correctness theorems for the WCNF-to-PB encoding.}\label{fig:enc-correct}
\end{figure}

\paragraph{Putting Everything Together.} The final verification step is to specialize the end-to-end machine code correctness theorem for \cakepb to the new frontend.
The resulting theorem for \cakepbwcnf is shown abridged in~\cref{fig:final-theorem}; a detailed explanation of similar \CakeML-based theorems is available elsewhere~\cite{GMMNOT24Subgraph,THM23VerifiedPropagation} so we do not go into details here.
Briefly, the theorem says that whenever the verdict string ``\texttt{s VERIFIED OUTPUT EQUIOPTIMAL}'' is printed (as a suffix) to the standard output by an execution of \cakepbwcnf, then the two input files given on the command line parsed to equioptimal MaxSAT WCNF instances.

\begin{figure}[t]
\begin{holthmenv}
\HOLTokenTurnstile{}\;\HOLConst{cake_pb_wcnf_run}\;\HOLFreeVar{cl}\;\HOLFreeVar{fs}\;\HOLFreeVar{mc}\;\HOLFreeVar{ms}\;\HOLSymConst{\HOLTokenImp{}}\\
\;\;\;\;\HOLSymConst{\HOLTokenExists{}}\HOLBoundVar{out}\;\HOLBoundVar{err}.\\
\;\;\;\;\;\;\HOLConst{extract_fs}\;\HOLFreeVar{fs}\;(\HOLConst{cake_pb_wcnf_io_events}\;\HOLFreeVar{cl}\;\HOLFreeVar{fs})\;\HOLSymConst{=}\\
\;\;\;\;\;\;\;\;\HOLConst{Some}\;(\HOLConst{add_stdout}\;(\HOLConst{add_stderr}\;\HOLFreeVar{fs}\;\HOLBoundVar{err})\;\HOLBoundVar{out})\;\HOLSymConst{\HOLTokenConj{}}\\
\;\;\;\;\;\;(\HOLConst{length}\;\HOLFreeVar{cl}\;\HOLSymConst{=}\;\HOLNumLit{4}\;\HOLSymConst{\HOLTokenConj{}}\;\HOLConst{isSuffix}\;\HOLStringLitDG{s VERIFIED OUTPUT EQUIOPTIMAL\HOLTokenBackslash{}n}\;\HOLBoundVar{out}\;\HOLSymConst{\HOLTokenImp{}}\\
\;\;\;\;\;\;\;\;\;\HOLSymConst{\HOLTokenExists{}}\HOLBoundVar{wfml}\;\ensuremath{\HOLBoundVar{wfml}\sp{\prime{}}}.\\
\;\;\;\;\;\;\;\;\;\;\;\HOLConst{get_fml}\;\HOLFreeVar{fs}\;(\HOLConst{el}\;\HOLNumLit{1}\;\HOLFreeVar{cl})\;\HOLSymConst{=}\;\HOLConst{Some}\;\HOLBoundVar{wfml}\;\HOLSymConst{\HOLTokenConj{}}\;\HOLConst{get_fml}\;\HOLFreeVar{fs}\;(\HOLConst{el}\;\HOLNumLit{3}\;\HOLFreeVar{cl})\;\HOLSymConst{=}\;\HOLConst{Some}\;\ensuremath{\HOLBoundVar{wfml}\sp{\prime{}}}\;\HOLSymConst{\HOLTokenConj{}}\\
\;\;\;\;\;\;\;\;\;\;\;\HOLConst{opt_cost}\;\HOLBoundVar{wfml}\;\HOLSymConst{=}\;\HOLConst{opt_cost}\;\ensuremath{\HOLBoundVar{wfml}\sp{\prime{}}})
\end{holthmenv}
\caption{Abridged final correctness theorem for \cakepbwcnf.}\label{fig:final-theorem}
\end{figure}

\section{Experiments}
\label{sec:experiments}


We upgraded the MaxSAT preprocessor \maxpre~2.1~\cite{IBJ22ClauseRedundancy,DBLP:conf/cp/JabsBIJ23,DBLP:conf/sat/KorhonenBSJ17} to \maxpre~2.2, which produces proof logs in the \veripb format~\cite{BMMNOT23DocumentationVeriPB}.
\maxpre~2.2 is available at the \maxpre~2 repository~\cite{MaxPre2}.
The generated proofs
were checked and elaborated using \veripb~\cite{VeriPB}
and the verified proof checker \cakepbwcnf.
As benchmarks 
we used the 558 weighted and 572 unweighted MaxSAT instances from the MaxSAT Evaluation 2023~\cite{MaxSATevaluation23}.

The experiments were conducted on 11th Gen Intel(R) Core(TM) i5-1145G7 @
2.60GHz CPUs with 16 GB of memory,
a solid state drive as storage, and running Rocky Linux 8.5.  
Each benchmark ran exclusively on a
node and the memory was limited to 14 GB. The time for \maxpre was limited to
300 seconds; there is also an option to let \maxpre know about this time limit, 
which we did not use since
\maxpre decides which techniques to try
based on the remaining time, which makes it difficult to compare
results
with and without proof logging.
The time limits for both  \veripb and
for \cakepbwcnf were set to 6000 seconds to get as many instances checked
as possible.

This evaluation focuses on the default setting of \maxpre, which does not use some of the techniques mentioned in~\cref{sec:logging-techniques}
\iflongversion
(or Appendix~\ref{apx:prepro}).
\else
(or the online appendix~\cite{}).
\fi
We have also conducted experiments with all techniques enabled to check the correctness of the proof logging implementation for all preprocessing techniques.
The data and source code from our experiments is available at~\cite{IOTBJMN24ExperimentalDataPreprocessing}.

The goal of the experimental evaluation is to answer the following questions:
\begin{description}
\item[\textbf{RQ1}.] How much extra time is required to write the proof for the preprocessor?
\item[\textbf{RQ2}.] How long does proof checking take compared to proof generation?
\end{description}

\begin{figure}[t]
  \begin{minipage}[c]{.45\textwidth}
    \includegraphics[width=\linewidth]{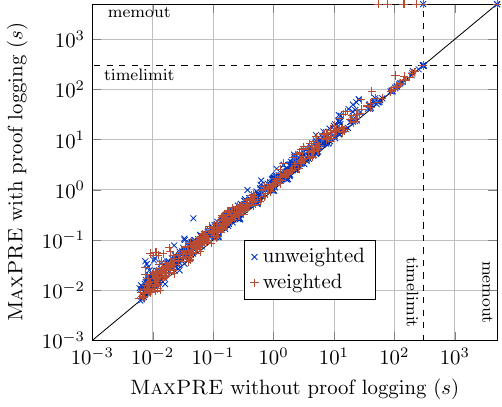}
    \caption{Proof logging overhead for \maxpre.}
    \label{fig:proof-logging-overhead}
  \end{minipage}
  \hfill
  \begin{minipage}[c]{.45\textwidth}
    \includegraphics[width=\linewidth]{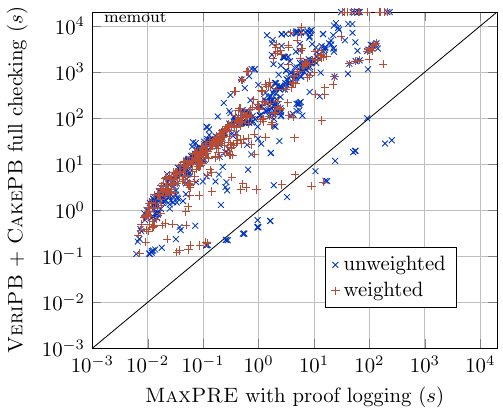}
    \caption{\maxpre vs. combined proof checking running time.}
    \label{fig:combined-checking}
  \end{minipage}
\end{figure}


To answer the first question,
we compare \maxpre with and without proof logging in~\cref{fig:proof-logging-overhead}.
In total, \maxpre without proof logging successfully preprocesses 1081 instances;
with proof logging enabled, 8 fewer instances were preprocessed within the resource limits, i.e., due to either time- or memory-out.
For the successfully preprocessed instances, the geometric mean of the proof logging overhead is 46\% of the running time and 95\% of the instances were preprocessed with proof logging in at most twice the time as without proof logging.

For our comparison between proof generation and proof checking,
we only consider the 1073 instances
successfully preprocessed by \maxpre within the resource limits.
Of these, 1021 instances were successfully checked and elaborated by
\veripb, and 991 instances were subsequently successfully checked by
the formally verified
checker \cakepbwcnf, with the remaining instances being time- or memory-outs.
This shows the practical viability of our approach, as the vast majority of preprocessing proofs were checked within the resource limits.

A scatter plot comparing the running time of \maxpre with proof logging enabled against the combined checking process is shown in Figure~\ref{fig:combined-checking}.
For the combined checking time, we only consider the instances that have been successfully checked by \cakepbwcnf. In the geometric mean, the time for the combined verified checking pipeline of \veripb elaboration followed by \cakepbwcnf is $113\times$ the preprocessing time of \maxpre.
A general reason for this overhead is that the preprocessor has more MaxSAT application-specific context than the pseudo-Boolean checker, so the preprocessor can log proof steps without performing the actual reasoning while the checker must ensure that those steps are sound in an application-agnostic way.
An example for this is reification: as the preprocessor knows its reification variables are fresh, it can easily emit redundance steps that witness on those variables; but the checker has to verify freshness against its own database.
Similar behaviour has been observed in other applications of pseudo-Boolean proof logging~\cite{GMNO22CertifiedCNFencodingPB,HOGN24CertifyingMIPpresolve}.

To analyse further the causes
of proof checking overhead, we also compared \veripb to \cakepbwcnf.
The checking of the elaborated kernel proof with \cakepbwcnf is $6.7\times$ faster than checking and elaborating the augmented proof with \veripb.
This suggests that the bottleneck for proof checking is \veripb; \veripb \emph{without} elaboration is about $5.3\times$ slower than \cakepbwcnf.
As elaboration is a necessary step before running \cakepbwcnf,
improving the
performance of \veripb would benefit the performance of the
pipeline as a whole.
One specific improvement would be to augment \rupname rules with
\lrat-style hints~\cite{CHHKS17EfficientCertified}, so that \veripb
does not need to elaborate \rupname to cutting planes.
This kind of
engineering challenges are
beyond the scope of this paper.

\section{Conclusion}
\label{sec:conclusion}

In this work, we show how to use pseudo-Boolean proof logging to
certify correctness of the MaxSAT preprocessing phase,
extending previous work for the main solving phase in unweighted
model-improving solvers~\cite{VWB22QMaxSATpb}
and general core-guided solvers~\cite{BBNOV23CertifiedCoreGuided}.
As a further strengthening of previous work, we present a fully
formally verified toolchain which provides end-to-end verification of
correctness.

In contrast to SAT solving, there is a rich variety of MaxSAT solving
techniques, and it still remains to design pseudo-Boolean proof
logging methods for general, weighted, model-improving MaxSAT solving
\cite{ES06TranslatingPB,LP10Sat4j,PRB18DynamicPolyWatchdog}
and
\emph{implicit hitting set (IHS)} solvers
\cite{DB11SolvingMAXSAT,DB13Exploiting}
with \emph{abstract cores}~\cite{BBP20AbstractCores}.
Nevertheless, our work demonstrates again that pseudo-Boolean reasoning
seems like a very promising foundation for MaxSAT proof logging, and
we are optimistic that this work is another step on the path towards
general adoption of proof logging techniques for MaxSAT solving.

%


\begin{credits}
  \subsubsection{\ackname}
  This work has been financially supported by the
  University of Helsinki Doctoral Programme in Computer Science DoCS,
  the Research Council of Finland under grants 342145 and 346056,
  the Swedish Research Council grants \mbox{2016-00782} and \mbox{2021-05165},
  the Independent Research Fund Denmark grant \mbox{9040-00389B},
  the Wallenberg AI, Autonomous Systems and Software Program (WASP)
  funded by the Knut and Alice Wallenberg Foundation,
  and by A*STAR, Singapore.
  Part of this work was carried out while some of the authors participated
  in the extended reunion of the semester program
  \emph{Satisfiability: Theory, Practice, and Beyond}
  in the spring of 2023 at the
  Simons Institute for the Theory of Computing at UC Berkeley.
  We also acknowledge useful discussions at the
  Dagstuhl workshops 22411
  \emph{Theory and Practice of SAT and Combinatorial Solving}
  and
  23261 \emph{SAT Encodings and Beyond}.
  The computational experiments were enabled by resources provided by
  LUNARC at Lund University.
\end{credits}

\iflongversion
\appendix
\section{Complete Overview of Proof Logging for MaxSAT Preprocessing}\label{apx:prepro}

In this appendix, we provide a complete overview of proof logging for the preprocessing techniques implemented by \maxpre.
As we already presented proof logging for bounded variable elimination, subsumed literal elimination,
label matching and binary core removal in Section~\ref{sec:logging-techniques} of the paper, we do not present those techniques here.
In addition, we do not include intrinsic at-most-ones (even though implemented in \maxpre), as it is already discussed in~\cite{BBNOV23CertifiedCoreGuided}.


\subsection{Fixing Variables}\label{sec:fix}
Many of the preprocessing techniques can fix variables (or literals) to either $0$ or $1$.
We describe here the generic procedure that is invoked when a variable is fixed.
Assume that a preprocessing technique decides to fix $\ell=1$ for a literal $\ell$.
Then, in the preprocessor, each clause $C\lor \olnot{\ell}$ is replaced by clause $C$, i.e., falsified literal $\olnot{\ell}$ is removed.
Additionally, each clause $C\lor \ell$ is removed (as they are satisfied when $\ell=1$).

In the proof, we do the following.
First, the technique that fixes $\ell=1$, ensures that constraint $\ell \ge 1$ is in the core constraints of the proof.
It may be that $\ell \ge 1$ is already in the core constraints of the proof (i.e.\ instance has a unit clause $(\ell)$),
or it may be that $\ell \ge 1$ needs to be introduced as a new constraint.
The details on how $\ell \ge 1$ is introduced depends on the specific technique that is fixing $\ell=1$.
Now, assuming $\ell \ge 1$ is in the core constraints, the following procedure is invoked.

\begin{enumerate}[(1)]
\item If $\ell$ or $\olnot{\ell}$ appears in the objective function, the objective function is updated.
\item For each clause $C\lor \olnot{\ell}$, the constraint $\asPB(C)$ is introduced as a sum of $\asPB(C\lor \olnot{\ell})$ and $\ell \ge 1$.
\item Each constraint $\asPB(C\lor \ell)$ is deleted (as a RUP constraint).
\item Finally, the core constraint $\ell \geq 1$ is deleted last with witness $\{\ell\rightarrow 1\}$.
\end{enumerate}

\subsection{Preprocessing on the Initial WCNF Representation} \label{sec:appwcnf}

We explain the preprocessing techniques that can be applied
during preprocessing on the WCNF representation, detailing especially how the different types
of clauses are handled.
The preprocessing techniques applied on the WCNF representation only modify a clause $C$ by
either removing a literal $\ell$ from $C$ or removing $C$ entirely.
With this intuition, given an input WCNF instance $\wcnfinstance = (\hards, \softs)$ and a working
 instance $\wcnfinstance_1 = (\hards^1, \softs^1)$
each clause in $\wcnfinstance_1$ is one of the following three types:
\begin{enumerate}[(1)]
\item A hard clause $C \in \hards^1$ that is a subset or equal to a hard clause $C \subseteq C^{\text{orig}} \in \hards$  of $\wcnfinstance$.
\item An \emph{originally unit soft clause}, i.e., a soft clause $C \in \softs^1$ that  is equal to a unit soft clause in $\softs$.
\item An \emph{originally non-unit soft clause}, i.e., $C \in \softs^1$ that is a subset or equal to a non-unit soft clause $C \subseteq C^{\text{orig}} \in \softs$ of $\wcnfinstance$.
\end{enumerate}

With this we next detail how the preprocessing rules permitted on the WCNF representation are logged. In the following, we assume a fixed
working WCNF instance.

\subsubsection{Duplicate Clause Removal.}
In the paper we discussed how to log the removal of two duplicate clauses $C$ and $D$ when:
\begin{inparaenum}[(i)]
\item both are hard, or
\item $C$ is hard and $D$ is an originally non-unit soft clause.
\end{inparaenum}
Here we detail the remaining cases.




Assume first that both $C$ and $D$ are originally non-unit duplicate soft clauses with weights $\weight^C$ and $\weight^D$, respectively.
Then the proof has the core constraints $\asPB(C\lor b_C)$ and $\asPB(D\lor b_D)$ and its objective the terms $\weight^C b_C$ and $\weight^D b_D$.
The removal of $D$ is logged as follows.
\begin{enumerate}[(1)]
\item \label{itm:dcr1} Introduce the constraints: $\olnot{b}_C+b_D\ge 1$ with the witness $\{b_C\rightarrow 0\}$ and $b_C+\olnot{b}_D\ge 1$, with the witness $\{b_D\rightarrow 0\}$ to the core set. These encode $b_C=b_D$.
\item Update the objective by adding $-\weight^{C}b_C+\weight^{C} b_D$ to it, conceptually increasing the coefficient of $b_D$ by $\weight^C$.
\item Remove the constraints introduced in step (\ref{itm:dcr1}) using the same witnesses.
\item Remove the (RUP) constraint $\asPB(D\lor b_C)$.
\end{enumerate}

If $C = (\olnot{\ell})$ is originally a unit soft clause but $D=(\olnot{\ell})$ is originally a non-unit soft clause,
then the core constraints of the proof include constraint
$\asPB(\ell \lor b_D)$ and the objective of the proof the terms  $\weight^C \ell$ and $\weight^D b_D$.
The removal of $D$ is logged similarly to the previous case with the literal $b_C$ replaced with $\ell$.

The case of two duplicate originally unit soft clauses does not require proof logging since the corresponding terms in
the objective are automatically summed.

\subsubsection{Tautology Removal.}
If a clause is a tautology, it is also a RUP clause.
Thus, a tautological hard clause is simply deleted.
The removal of a tautological soft clause additionally requires
updating the objective.

More specifically, assume $C$ is a tautological soft clause of weight $\weight^C$.  Then $C$
is originally non-unit, so the proof has a constraint $\asPB(C \lor b_C)$ and its objective the
term $\weight^C b_C$.
The removal of $C$ is logged with the following steps:
\begin{enumerate}[(1)]
\item Delete the (RUP) constraint $\asPB(C\lor b_C)$.
\item \label{itm:tauto2} Introduce the constraint $\olnot{b}_C\ge 1$ with witness $\{b_C \rightarrow 0\}$ and move the new constraint to the core set.
\item Update the objective by adding $-\weight^C b_C$ to it.
\item Remove the constraint introduced in step (\ref{itm:tauto2}) with the same witness.
\end{enumerate}

\subsubsection{Unit Propagation of Hard Clauses.}
If the instance contains a (hard) unit clause $(l)$, the literal $l$ is fixed to $1$ with the method of fixing variables described in Section~\ref{sec:fix}.

\subsubsection{Removal of Empty Soft Clauses.}
If the instance contains an empty soft clause $C$---either as input or as a consequence of e.g.,\ unit propagation---it is removed and the
lower bound increased by its weight $\weight^C$.
If $C$ was originally non-unit, the core constraints of the proof contain the constraint $b_C \ge 1$ and the objective the term
$\weight^C b_C$. The removal of $C$ is logged by the following steps:
\begin{enumerate}[(1)]
\item Update the objective by adding $-\weight^C b_C+\weight^C$.
\item Delete the constraint $b_C \ge 1$ with the witness $\{b_C \to 1\}$.
\end{enumerate}

If $C = (\ell)$ is an originally unit soft clause the objective is updated in conjunction with the literal $\ell$ getting fixed to $0$, as
described in Section~\ref{sec:fix}. Thus, no further steps are required.

\subsubsection{Blocked Clause Elimination (BCE)~\cite{DBLP:conf/tacas/JarvisaloBH10}.}
Our implementation of BCE considers a clause $C\lor \ell$ blocked (on the literal $\ell$)
if for each clause $D\lor \olnot{\ell}$ there is a literal
$\ell'\in D$ for which $\olnot{\ell'}\in C$.

When preprocessing on the objective-centric representation, BCE considers only literals $\ell$ for which neither $\ell$ nor $\olnot{\ell}$ appears
in the objective function.
During initial WCNF preprocessing stage, there are no requirements for literal $\ell$.
(Notice that whenever there is a unit clause $(\olnot{\ell})$, $C\lor \ell$ is not blocked on the literal $\ell$.)

The removal of a blocked clause is logged as the deletion of the corresponding constraint
$\asPB(C\lor \ell)$ with the witness $\{\ell\rightarrow 1\}$.
If $C\lor \ell$ is an (originally non-unit) soft clause, the objective function is also updated exactly as with tautology removal.

\subsubsection{Subsumption Elimination.}
A clause $D$ is subsumed by the clause $C$ if $C\subseteq D$.
Whenever the subsuming clause $C$ is hard, $D$ is removed as
a RUP clause. If $D$ is soft, the objective function is updated exactly as with tautology removal.

\subsection{Preprocessing on Objective-Centric Representation}

We detail how the preprocessing techniques that are applied on the objective-centric representation $(\formula, \obj)$ of the working instance
are logged. In addition to these, the preprocessor can also apply the techniques  detailed in Section~\ref{sec:appwcnf}.

\subsubsection{TrimMaxSAT~\cite{DBLP:conf/vmcai/PaxianRB21}.}
The TrimMaxSAT technique heuristically looks for a set of literals $N$ s.t. every solution $\rho$ to $\formula$ assigns each $\ell \in N$ to $0$, or more formally,
$\formula$ entails the unit clause $(\olnot{\ell})$.
All such literals are fixed by the generic procedure (recall Section~\ref{sec:fix}).
The literals to be fixed are identified by iterative calls to an (incremental) SAT solver~\cite{DBLP:journals/entcs/EenS03,DBLP:series/faia/0001LM21} under different assumptions.

In order to log the TrimMaxSAT technique we log the proof produced by each SAT solver call into the derived set of constraints in our PB proof.
After the set $N$ is identified, we make $|N|$ extra SAT calls, one for each $\ell \in N$. Each call is made assuming the value of $\ell$ to $1$.
Due to the properties of TrimMaxSAT and SAT-solvers, the result will be UNSAT, after which $\olnot{\ell} \geq 1$ will be RUP w.r.t to the current set of core and derived
constraints. As such it is added and moved to core in order to invoke the generic variable fixing procedure.
Finally, when TrimMaxSAT will not be used any more,
all constraints added to the derived set by the SAT solver are removed.

\subsubsection{Self-Subsuming Resolution (SSR)~\cite{EB05EffectivePreprocessing,DBLP:conf/cp/OstrowskiGMS02}.}
Given clauses $C\lor l$ and $D\lor \olnot{\ell}$ such that $C$ subsumes $D$ and $\ell$ is not in the objective, SSR substitutes $D$ for $D\lor \olnot{\ell}$.
The proof has two steps:
\begin{inparaenum}[(1)]
\item Introduce $\asPB(D)$ as a new RUP constraint.
\item Remove $\asPB(D\lor \olnot{\ell})$ as it is RUP.
\end{inparaenum}

\subsubsection{Group-Subsumed Label Elimination (GSLE)~\cite{DBLP:conf/sat/KorhonenBSJ17}.}
Let $b$ be an objective variable that has the coefficient $c^b$ in $\obj$,  and $L$ a set of objective variables such that
each $b_i \in L$ has coefficient $c^i$ in $\obj$.
Assume then that (i) $c^b \geq \sum_{b_i\in L} c^i$, (ii) the negation of $b$ or any
variables in $L$ do not appear in any clauses, and
(iii) $\{C\mid b\in C\}\subseteq\{D\mid \exists b\in L : b\in D\}$.
Then, an application of GSLE
fixes $b=0$.
To prove an application of GSLE, we introduce the constraint $\olnot{b}\ge 1$ with the witness  $\{b\rightarrow 0, b_i\rightarrow 1 \mid b_i \in L\}$,
and invoke the generic variable fixing procedure detailed in Section~\ref{sec:fix} to fix $b=0$.

\subsubsection{Bounded Variable Addition (BVA)~\cite{MHB12AutomatedReencoding}.}

Consider a set of literals $M_{\mathit{lit}}$ and a set of clauses $M_{\mathit{cls}}\subseteq \formula$, such that for all $\ell\in M_{\mathit{lit}}$ and $C\in M_{\mathit{cls}}$,
each clause $(C\setminus M_{\mathit{lit}} \cup \{\ell\})$ is either in $\formula$ or a tautology.
Then an application of BVA adds the clauses $S_{x}=\{(\ell \lor x)\mid l\in M_{\mathit{lit}}\}$ and $S_{\olnot{x}}=\{(C\setminus M_{\mathit{lit}})\cup \{\olnot{x}\}\mid C\in M_{\mathit{cls}}\}$,
and removes the clauses $C\setminus M_{\mathit{lit}}$.

An application of BVA is logged as follows:
\begin{inparaenum}[(1)]
\item Add the constraint $\asPB(C)$ for each $C\in S_{\olnot{x}}$ with the witness $\{x\rightarrow 0\}$.
\item Add the constraint $\asPB(C)$ for each $C\in S_{x}$ with the witness $\{x\rightarrow 1\}$.
\item Delete each constraint $\asPB(C)$ for $C\in M_{\mathit{cls}}$ as a RUP constraint.
\end{inparaenum}

\subsubsection{Structure-based Labelling~\cite{DBLP:conf/sat/KorhonenBSJ17}.}
Given an objective variable $b$ and a clause $C$ that is blocked  on the literal $\ell$, when $b=1$,
an application of structure-based labelling replaces $C$ with $C\lor b$.
The proof is logged as follows:
\begin{inparaenum}[(1)]
\item Introduce the constraint $\asPB(C\lor b)$ that is RUP.
\item Delete the constraint $\asPB(C)$ with the witness $\{\ell\rightarrow 1\}$.
\end{inparaenum}

\subsubsection{Failed Literal Elimination (FLE)~\cite{Freeman1995ImprovementsPropositionalSatisfiability,DBLP:journals/endm/Berre01,DBLP:conf/aaai/ZabihM88}.} 
A literal $\ell$ is failed (denoted  $\ell \up\confl$)
if setting $\ell=1$ allows unit propagation to derive a conflict (i.e., an empty clause).
An application of FLE fixes $\ell=0$ when $\ell$ is a failed literal
for which $\olnot{\ell}$ is not
in the objective.

In addition to standard FLE, \maxpre implements an
extension that also fixes a literal $\ell = 0$ if:
\begin{inparaenum}[(i)]
\item $\olnot{\ell}$ is not in the objective function
\item each clause in $\formula$ that contains
$\ell$ also contains some other literal $\ell'$ that is implied by $\ell$ by unit propagation (denoted $\ell \up \ell'$), i.e.,
setting $\ell=1$ also fixes $\ell'=1$ after a sequence of unit propagation steps is applied.
\end{inparaenum}


\paragraph{Logging FLE.} For a failed literal $\ell$ the constraint $\olnot{\ell} \geq 1$ is RUP.
For the extended technique the constraint $\olnot{\ell} \geq 1$ is introduced
with the witness $\{\ell \rightarrow 0\}$.
Afterwards the generic procedure for fixing literals described in Section~\ref{sec:fix} is invoked.

\subsubsection{Implied Literal Detection.}
If both a literal $\ell_1$ and its negation $\olnot{\ell}_1$ imply another literal $\ell_2$ by unit propagation
(i.e., propagating either $\ell=1$ or $\ell=0$ also propagates $\ell_2=1$), the preprocessor fixes $\ell_2=1$.

As an extension to this technique, the preprocessor also fixes $\ell_2=1$ if
\begin{inparaenum}[(i)]
\item $\ell_1$ implies $\ell_2$ by unit propagation,
\item neither $\ell_1$ nor $\ell_2$ appear in the objective function in either polarity, and
\item each clause containing $\olnot{\ell}_2$ also contains some other literal $\ell'$ that is implied by $\olnot{\ell}_1$ by unit propagation.
\end{inparaenum}

\paragraph{Logging Implied Literals.}
For some intuition, note that $\ell_1\up \ell_2$ does not in general imply $\olnot{\ell}_2\up \olnot{\ell}_1$.
Thus, there is no guarantee that $\ell_2 \geq 1$ would be RUP.
Given that $\ell_1 \up \ell_2$ and $\olnot{\ell}_1\up \ell_2$, the proof is instead logged as follows:
\begin{enumerate}[(1)]
\item \label{itm:implit1} Add $\olnot{\ell}_1 + \ell_2 \geq 1$ and $\ell_1 + \ell_2 \geq 1$ that are both RUP.
\item Introduce the constraint $\ell_2 \geq 1$ by divide the sum of constraints introduced in step (\ref{itm:implit1}) by $2$. Move the new constraint to the core constraints.
\item Delete the constraints introduced in step (\ref{itm:implit1}).
\item Invoke the generic procedure detailed in Section~\ref{sec:fix} to fix $\ell_2=1$.
\end{enumerate}

The extended technique is logged by first adding the constraint $\ell_1 + \ell_2 \geq 1$ with the witness $\{\ell_2 \rightarrow 1\}$.
For some intuition, if the constraint is falsified, the assumptions
guarantee that $\ell'=1$ so the value of $\ell_2$ can be flipped without falsifying other constraints.

\subsubsection{Equivalent Literal Substitution~\cite{DBLP:journals/tsmc/Brafman04,DBLP:conf/aaai/Li00,DBLP:journals/amai/Gelder05}.}
If $\ell_1 \up \ell_2$ and $\olnot{\ell}_1\up \olnot{\ell}_2$, the equivalent literal technique substitutes $\ell_1$ with $\ell_2$.
As an extension to this technique, the same substitution is applied also in cases where the following three conditions hold:
\begin{inparaenum}[(i)]
\item $\ell_1 \up \ell_2$,
\item neither $\ell_1$ nor $\ell_2$ appear in the objective function in either polarity, and
\item $\olnot{\ell}_1$ implies some other literal in
each clause containing $\olnot{\ell}_2$ by unit propagation.
\end{inparaenum}

\paragraph{Logging Equivalent Literals.}
An application of equivalent literal substitution is logged as follows.
\begin{enumerate}[(1)]
\item\label{itm:equivlit1} Introduce the clauses $\olnot{\ell}_1 + \ell_2 \geq 1$ and $\ell_1 + \olnot{\ell}_2 \geq 1$ as RUP. In the case of the extended technique, $\ell_1 + \olnot{\ell}_2 \geq 1$ is added with the witness $\{\ell_2 \rightarrow 0\}$.
\item For each clause $C\lor \ell_1$, replace $\asPB(C\lor \ell_1)$ with $\asPB(C\lor \ell_2)$ with the RUP rule.
\item For each clause $C\lor \olnot{\ell}_1$, replace $\asPB(C\lor \olnot{\ell}_1)$ with $\asPB(C\lor \olnot{\ell}_2)$ with the RUP rule.
\item If $\ell_1$ or $\olnot{\ell}_1$ appear in the objective function, replace them with $\ell_2$ and $\olnot{\ell}_2$, respectively.
\item Remove the constraints introduced in step (\ref{itm:equivlit1}).
\end{enumerate}

\subsubsection{Hardening~\cite{DBLP:conf/cp/AnsoteguiBGL12,IBJ22ClauseRedundancy,DBLP:conf/sat/MorgadoHM12}.}
Given an upper bound $UB$ for the optimal cost of $(\formula, \obj)$ and an objective variable $b$ that has a coefficient $w^{b}>UB$ in $\obj$,
hardening fixes $b=0$.
Proof logging for hardening has been previously studied in~\cite{BBNOV23CertifiedCoreGuided}.
In~\cite{BBNOV23CertifiedCoreGuided}, however, the hardening is done with the presence of so-called objective-improving constraints, i.e., constraints of form $\obj\le UB-1$, where $UB$ is the cost of the best currently known solution.
In the context of preprocessing where the preprocessor should provide an equioptimal instance as an output, introducing objective-improving constraints to the instance is not possible.
Instead, given a solution $\rho$ to $\formula$ with cost $\obj(\rho)=UB$ and an objective variable $b$ with $w^{b}>UB$,
we introduce the constraint $\olnot{b} \geq 1$ with $\rho$ as the witness
and then invoke the generic procedure for fixing variables, as detailed in Section~\ref{sec:fix}.

\subsection{Conversion to WCNF --- Renaming Variables}
In the final stage of preprocessing, \maxpre converts the instance to WCNF.
The conversion removes the objective constant as described in Section~\ref{sec:mprepro5} of the main paper.
Additionally,  the conversion `renames' (some of) the variables.

There are two reasons for renaming variables. The first is to remove any gaps in the indexing of variables.
In WCNF, variables are named with integers.
During preprocessing, some variables in the instance might have been eliminated from the instance. At the end \maxpre
compacts the range of variables to be continuous and start from $1$.
The second reason for renaming variables is to sync names between WCNF and the pseudo-Boolean proof.
In the pseudo-Boolean proofs, the naming scheme of variables is different,
valid variable names include, for instance, \texttt{x1}, \texttt{x2}, \texttt{y15}, \texttt{\_b4}.
When a WCNF instance is converted to a pseudo-Boolean instance, the variable \texttt{i} of the WCNF instance is mapped to the variable \texttt{xi}
of the pseudo-Boolean instance.
For $j$th non-unit soft clause of a WCNF instance, the conversion introduces a variable \texttt{\_bj}.
During preprocessing, the `proof logger' of \maxpre takes care of mapping \maxpre variables to correct variable names in proof.
In the end, however, \maxpre produces an output WCNF file, and at this point, each variable \texttt{i} of WCNF instance should again correspond to variable
\texttt{xi} of proof.
Thus, for example, all \texttt{\_b}-variables are replaced with \texttt{x}-variables.

\paragraph{Logging variable naming.}
Assume that the instance has a set of variables $V$ and for each $x\in V$, we wish to use name $f(x)$ instead of $x$ in the end.
We do proof logging for variable renaming in two phases.
\begin{inparaenum}[(1)]
\item For each $x\in V$, introduce temporary variable $t_x$, set $x=t_x$ and then `move' all the constraints and the objective function to the temporary namespace.
The original constraints and encodings for $x=t_x$ are then removed.
\item For each $x\in V$, introduce $f(x)=t_x$, and `move' the constraints and the objective to the final namespace.
The temporary constraints and encodings are then removed.
\end{inparaenum}

\subsection{On Solution Reconstruction and Instances Solved During Preprocessing}\label{apx:nosols}
Finally, we note that while the focus of this work has been on
certifying the preservation of the costs of solutions, in practice
our certified preprocessor also allows reconstructing a minimum-cost solution
to the input. More precisely, consider an input WCNF instance $\wcnfinstance$,
a preprocessed instance $\wcnfinstance_P$, and an optimal solution
$\rho_p$ to $\wcnfinstance_P$. Then \maxpre can compute an optimal
solution $\rho$ to $\wcnfinstance$ in linear time with respect to the number of preprocessing
steps performed. More details can be found in~\cite{DBLP:conf/sat/KorhonenBSJ17}.

Importantly, the optimality of a reconstructed solution can be easily verified without considering how the reconstruction is implemented in practice;
given that we have verified the equioptimality of $\wcnfinstance$ and $\wcnfinstance_P$, and that $\rho_p$ is an optimal solution to $\wcnfinstance_P$,
the optimality of reconstructed $\rho$ to $\wcnfinstance$ can be verified by checking that
\begin{inparaenum}[(i)]
\item $\rho$ indeed is a solution to $\wcnfinstance$
\item The cost of $\rho$ w.r.t. $\wcnfinstance$ is equivalent to the cost of $\rho_p$ w.r.t. $\wcnfinstance_P$.
\end{inparaenum}

On a related note, \maxpre can actually solve some instances during preprocessing, either
by: (i) determining that the hard clauses do not have solutions, or (ii) computing an optimal solution
to some working instance. In practice (i) happens by the derivation of the unsatisfiable empty (hard) clause
and (ii) by the removal of every single clause from the working instance.
We have designed the preprocessor to always terminate with an output
WCNF and a proof of equioptimality rather than producing different kinds of proofs.

If an empty hard clause is derived, the preprocessing is immediately terminated
and an output WCNF instance containing a single hard empty clause produced.
Additionally, an empty constraint $0 \geq 1$ is added to the proof and
all other core constraints deleted by the RUP rule. Notice how the proof
of equioptimality between the input and output can in this case be seen as a proof of infeasibility
of the input hard clauses.

If all clauses are removed from the working instance,
MaxPRE terminates and outputs the instance obtained after constant
removal (recall Stage 5 in Section~\ref{sec:logging-techniques}) on an instance without
other clauses.

\fi

%
%
\bibliographystyle{splncs04}

\bibliography{paper,refArticles.bib,refBooks.bib,refOther.bib}

\end{document}